\documentclass[smallextended]{svjour3}       

\smartqed  
\PassOptionsToPackage{table,xcdraw,dvipsnames, usenames}{xcolor}
\usepackage{graphicx}
\usepackage{subcaption}
\usepackage{url}
\usepackage{enumitem}
\usepackage{tcolorbox}

\usepackage{multirow}
\usepackage{amsmath}
\usepackage{array,multirow}
\usepackage{adjustbox}
\usepackage{booktabs}
\usepackage{soul}
\usepackage[english]{babel}
\usepackage{anyfontsize}
\usepackage{algorithm}
\usepackage{float}
\usepackage[]{xcolor}

\usepackage[normalem]{ulem}
\useunder{\uline}{\ul}{}

\graphicspath{{figures/}} 

\usepackage{todonotes}
\newcommand{\sectopic}[1]{\vspace{0em}\par\noindent{\textit{\bfseries #1}}}

\usepackage{varwidth}
\usepackage{ragged2e}
\usetikzlibrary{calc}

\newcounter{tipcounter}
\usepackage{tikz}

\newcommand{\sys}{\textsf{RoboREIT}}
\newcommand{\vico}{\textsf{VICO}}

\begin{document}

\title{RoboREIT: an Interactive Robotic Tutor with Instructive Feedback Component for Requirements Elicitation Interview Training}


\author{Binnur Görer         \and
        Fatma Başak Aydemir 
}
\institute{Binnur Görer \at
              Boğaziçi University, Istanbul, Turkey \\
              \email{binnur.gorer@boun.edu.tr}           
           \and
           Fatma Başak Aydemir \at
              Boğaziçi University, Istanbul, Turkey \\
              \email{basak.aydemir@boun.edu.tr}
}

\date{Received: date / Accepted: date}

\maketitle
\begin{abstract}
\textbf{[Context]} Requirements elicitation activities include actively seeking, uncovering, acquiring, and processing the needs and desires of the stakeholders. Interviewing stakeholders is the most popular requirements elicitation technique among multiple techniques. The success of an interview depends on the collaboration of the interviewee which can be fostered through the interviewer's preparedness and communication skills. Mastering these skills requires experience and practice interviews provide students the chance to apply their theoretical knowledge and gain experience. 
\textbf{[Problem]} 
This type of practical training is resource-heavy as it calls for the time and effort of a stakeholder for each student which may not be feasible for a large number of students. 
\textbf{[Principal Idea]} This paper proposes \sys{}, an interactive Robotic tutor for Requirements Elicitation Interview Training to provide a means for students to gain experience and a solution to the scalability problem of the elicitation interviews training. The humanoid robotic component of \sys{} responds to the questions of the interviewer, which the interviewer chooses from a set of predefined alternatives for a particular scenario. After the interview session, \sys{} provides contextual feedback to the interviewer on their performance and allows the student to inspect their mistakes. \sys{} is extensible with various scenarios.
\textbf{[Results]} We performed an exploratory user study to evaluate \sys{} and demonstrate its applicability in requirements elicitation interview training. The quantitative and qualitative analyses of the users' responses reveal the appreciation of \sys{} and provide further suggestions about how to improve it. 
\textbf{[Contribution]} Our study is the first in the literature that utilizes a social robot in requirements elicitation interview education. \sys{}'s innovative design incorporates replaying faulty interview stages and allows the student to learn from mistakes by a second time practicing. All participants praised the feedback component, which is not present in the state-of-the-art, for being helpful in identifying the mistakes. A favorable response rate of 81\% for the system's usefulness indicates the positive perception of the participants.

\end{abstract}

\section{Introduction}\label{sec:introduction}

The purpose of requirements elicitation is to seek, uncover, acquire and process the needs and desires of the stakeholders~\cite{zowghi2005requirements}. Requirements elicitation is the initial stage for software development, and its success affects how well the subsequent stages and the entire project turn out ultimately~\cite{van2009requirements,chen2019contemporary}. A software development project could fail if the system requirements are not gathered in a thorough, consistent, and unambiguous manner~\cite{shafiq2018effect}. Among the many elicitation procedures available, requirements elicitation interviews are considered the most common and efficient method for they often allow for more thorough information gathering through direct communication with the stakeholder in a limited time period~\cite{davis2006effectiveness}. Since the stakeholder side is typically untrained in conducting elicitation interviews, the requirements engineer is expected to plan and manage the interview process. A successful software product project should therefore have competent requirements engineers who thrive in conducting interviews.

Elicitation interviews require various skills, including the ability to establish a rapport with the interviewee to make the process go more smoothly, formulate questions correctly, and assess the stakeholders' perspectives and requirements. Interpersonal interaction is essential when conducting interviews since effective communication may help significantly bridge the gap between requirements engineers and stakeholders due in part to a lack of shared domain expertise~\cite{hadar2014role}. However, it is difficult to improve these skills, and traditional textbook-centered training approaches are insufficient unless they are complemented with practice. Along with the theoretical understanding of the interview techniques, constant practice in a setting where real interview dynamics are held is essential for strengthening the skills like interview management, behavioral control, and looking confident. Practicing can also help students to alleviate the well-known phenomena of interview nervousness~\cite{andrews2006student,powell2021shake}. 

Role-playing is widely adopted in teaching requirements elicitation interviews in order to address the need for practicing. It can be carried out in a variety of ways, such as by putting students in pairs or with teaching assistants or instructors. In multiple interview exercises, the roles of the requirements engineer and stakeholder might be switched around to let the student experience the roles and expectations of both parties. However, due to the excessive amount of human effort needed to plan and oversee the interviews, these activities are not always feasible to apply in a regular classroom context. Even if applied, the activity's duration and repetition would be constrained, which would not allow the students to practice enough to enhance their interviewing skills. Moreover, it does not guarantee that every student will have an equally good experience. Some students may be paired with someone who excels in their role, while others may be paired with someone who has little interest in the activity and performs poorly~\cite{debnath2020designing}. Technological tools like games, domain expert systems, and simulations can overcome these limitations of human-human role-playing activities by incorporating a digital interview partner~\cite{daun2021survey}.

Social robots are becoming a more prevalent source of help in our daily lives, particularly in professions like teaching, training, and education, where success depends on regular practice and attentive supervision. Their interactive appearance and expressions give them a sense of presence that disembodied or virtual entities may lack~\cite{belpaeme2018social,mubin2018social}. Thanks to their many capabilities, robots are evolving into valuable components of the educational ecosystem. Robots can benefit education in various ways, including interactive learning, helping to practice, and social skill development. They can provide the same level of service to every learner without having any human prejudices based on factors like gender, ethnicity, or age. Moreover, students can access a robotic trainer at any time from anywhere and can repeat the training any number of times they need. Robots do not get fatigued like human instructors do, especially for repetitive jobs requiring a high cognitive load, such as practicing conducting interviews. By using robots for this kind of duty, instructors could be freed up to concentrate more on creating new pedagogical strategies or improving the course contents. Even though the social robot systems available today cannot yet meet the high expectations of education experts who emphasize the need for more complex systems with high cognitive and interactive skills \cite{sonderegger2022social}, there are several pilot studies that prove the value of using social robots in various teaching scenarios. In this work, we focus on one potential role: an interview trainer, which can help novice requirements engineers and students to excel in conducting elicitation interviews before having actual interviews.

We present \sys{}, an interactive Robotic tutor for Requirements Elicitation Interview Training that includes a fully worked out interaction scheme for practicing elicitation interviews as a stakeholder and a feedback utility to monitor and replay the mistakes of the student. We assessed the user experience and the applicability of the system in a real class setting of a graduate-level requirements engineering course. Following the online learning trends that emerged during the Covid-19 pandemic, we build our system to be used remotely via a video conferencing tool at any time, from anywhere, without requiring the physical presence of the user with the robot (a Nao robot is used), though \sys{} can function in person as well if preferred. The system is designed to function autonomously (except for speech to text component) to minimize human intervention. 

The proposed system operates in two main phases. In the interview phase, the student plays the role of a requirements engineer and \sys{} acts as a stakeholder for the project. Following a predefined scenario, at each turn of the interview, \sys{} provides multiple choices for the interviewer’s next question. The scenario unfolds based on the responses (selection of questions) of the interviewer. After the interview process, \sys{} functions as a tutor by providing feedback on the interview performance of the student to improve their interview skills. It tracks the technical mistakes of the student, allows them to revisit the incorrect parts, and practices again by reinforcing appropriate contextual feedback. Finally, \sys{} wraps up and ends the training session after the interviewer reviews all the incorrect responses. Currently, \sys{} utilizes the scenario presented in~\cite{debnath2020designing}. However, \sys{} is designed to work with any other scenario for another domain or project, or including a specific subset of types of interview mistakes identified by Bano \emph{et al.}~\cite{bano2019teaching}.

To the best of our knowledge, \sys{} is the first interactive system that utilizes a social robot for requirements elicitation interview training. We evaluated \sys{} with a diverse group of software engineering professionals and students in comparison to \vico{}, a web-based interview simulator proposed by Debnath and Spoletini~\cite{debnath2020designing}. We quantitatively evaluated \sys{} by inspecting the participants’ interactions with the system and their responses to a series of questionnaires. During interview turns, the participants reacted noticeably more quickly in \sys{} than in \vico{}, which may be attributed to the audio-visual interaction modality offered by an embodied social robot. \sys{} may have encouraged the participants to speak as if they were having a conversation with another human. This supports our design goal of developing a realistic interview setup. 

The evaluation results are promising. The participants rated \sys{} favorably (i.e., higher than 3 = moderate level) for the perceived acceptance, engagement, and helpfulness in spotting the mistakes.  Yet, we do not observe a significant difference with the corresponding ratings of \vico{}. In terms of perceived ease-of-use, \vico{} scored higher than \sys{}, which makes sense given its straightforward construction: While our system forces the participant to listen to the robot actively and verbally answering by the selected option, \vico{} features a web-based interface that allows the user to conduct the interview by reading the texts on the screen and replying with a mouse click. The perceived ease-of-use may also be affected by the complexity of our feedback component. The feedback component of \sys{} is more demanding because it requires revisiting each incorrect interview turn by obtaining feedback and performing a second examination. Still, our findings confirmed that the users highly appreciated \sys{} as an interview trainer despite its elaborate form of interaction and compelling feedback utility, both of which were designed with the intention of providing the user with a realistic interview setting and the opportunity to inspect, correct, and learn from their own mistakes. To make our system and experimental produce reproducible, the implementation of \sys{}, the supporting materials for the experiment i.e., the introductory video, pre-experiment survey, post-experiment survey, are made available~\cite{binnur_gorer_2022_7263541}.

The remainder of the paper is structured as follows. Section~\ref{sec:related_work} presents the related work for the innovative approaches in requirements engineering (RE) education and the role of social robots in the education domain. In Section~\ref{sec:system}, we describe \sys{}'s system architecture and flow of interaction with the interviewer. In Section~\ref{sec:evaluation}, we present our research questions, the user study design, and the results of the comparative evaluation of our system. Section~\ref{sec:threats_to_validity} explains the threats to validity and how we mitigate them. Section~\ref{sec:conclusion} outlines the discussion points, the limitations of the study, and the concluding remarks.

\section{Related Work}
\label{sec:related_work}
In requirements engineering education and training (REET), it is essential to combine academic knowledge with practical application to give students experiences relevant to real industry cases~\cite{daun2021survey}. Conventional pedagogical approaches require instructors or peers to actively participate in the practical training, which limits the repeatability and length of the project. Although technological advancements could help address this issue, not many studies utilize new technologies in the field of REET. In this section, we briefly outline the existing studies in the literature in terms of the technology used, the requirements elicitation steps targeted in training, and the validation methodology of the suggested approach. Following that, the application of social robots, as one of the most prominent technologies, in a broader range of educational settings is examined in light of the available research. 

\begin{table*}
\caption{Summary of the requirement engineering education and training research which use new technologies.}
\label{tab:related_work}
\scriptsize
\begin{tabular}{llcll}
\hline \\[-0.5em]
\rowcolor[HTML]{FFFFFF} 
Reference                                                                                      & Year & Technology                                                                                          & \begin{tabular}[c]{@{}c@{}}Focused RE \\ Steps\end{tabular}                     & \begin{tabular}[c]{@{}c@{}}\# participants \\ in user study\end{tabular} \\ \\[-0.5em] \hline \\[-0.5em]
Garbers \textit{et. al.} \cite{garbers2006light}          & 2006 & \begin{tabular}[c]{@{}c@{}}interactive document\\  editing tool for \\ SRS preparation\end{tabular} & \begin{tabular}[l]{@{}l@{}}specification,\\ validation\end{tabular}             & Not mentioned                                                            \\ \\[-0.5em]
Vega \textit{et. al.} \cite{vega2009training}              & 2009 & serious game                                                                                        & elicitation                                                                     & Not applicable                                                           \\ \\[-0.5em]
Liang \textit{et. al.} \cite{liang2010experiences}         & 2010 & \begin{tabular}[c]{@{}c@{}}wiki tool for SRS \\ preparation\end{tabular}                            & \begin{tabular}[l]{@{}l@{}}elicitation, \\ specification\end{tabular}           & 26 students                                                              \\ \\[-0.5em]
Hainey \textit{et. al.} \cite{hainey2011evaluation}        & 2011 & serious game                                                                                        & \begin{tabular}[l]{@{}l@{}}elicitation, \\ analysis\end{tabular}                & 174 students                                                             \\ \\[-0.5em]
Ogata \textit{et. al.} \cite{ogata2012training}            & 2012 & \begin{tabular}[c]{@{}c@{}}support tool for \\ automatic protype \\ generation\end{tabular}            & specification                                                                   & 35 students                                                              \\ \\[-0.5em]
Nakamura \textit{et. al.} \cite{nakamura2014requirements}  & 2014 & domain expert system                                                                                & \begin{tabular}[l]{@{}l@{}}elicitation, \\ analysis\end{tabular}                & 10 students                                                              \\ \\[-0.5em]
Kakeshita \textit{et. al.} \cite{kakeshita2015requirement} & 2015 & assistive SW tool                                                                                   & elicitation                                                                     & 5 grad students                                                          \\ \\[-0.5em]
Yasin \textit{et. al.} \cite{yasin2018design}              & 2018 & serious game                                                                                        & \begin{tabular}[l]{@{}l@{}}elicitation, \\ analysis, \\ validation\end{tabular} & 16 students                                                              \\ \\[-0.5em]
Paschoal \textit{et. al.} \cite{paschoal2018chatterbot}    & 2018 & chatbot                                                                                             & elicitation                                                                     & 19 students                                                              \\ \\[-0.5em]
Garcia \textit{et. al.} \cite{garcia2019experiences}       & 2019 & serious game                                                                                        & elicitation                                                                     & 94 students                                                              \\ \\[-0.5em]
Ochoa \textit{et. al.} \cite{ochoa2019incorporating}       &  2019 & virtual reality experience                                                                          & elicitation                                                                     & Not mentioned                                                            \\ \\[-0.5em]
Ibrahim \textit{et. al.} \cite{ibrahim2019design}          & 2019 & serious game                                                                                        & \begin{tabular}[l]{@{}l@{}}elicitation, \\ analysis\end{tabular}                & 10 students                                                              \\ \\[-0.5em]
Garcia \textit{et. al.} \cite{garcia2020serious}           & 2020 & serious game                                                                                        & \begin{tabular}[l]{@{}l@{}}elicitation, \\ analysis\end{tabular}                & 80 students                                                              \\ \\[-0.5em]
Laiq \textit{et. al.} \cite{laiq2020chatbot}               & 2020 & chatbot                                                                                             & elicitation                                                                     & Not applicable                                                           \\ \\[-0.5em]
Debnath \textit{et. al.} \cite{debnath2020designing}       & 2020 & interview simulator                                                                                 & elicitation                                                                     & 17 students                                                              \\ \\[-0.5em] \hline
\end{tabular}
\end{table*}

\subsection{Use of Technologies in Requirements Engineering Education}
\label{sec:background_re_edu}
We provide an up-to-date investigation of the REET studies which use a certain technology as an education tool in teaching or training a set of RE steps. We summarized them in Table~\ref{tab:related_work} by mentioning the implemented technology, focused RE steps, and the number of participants the proposed approach is evaluated with. We only included the studies which implemented or demonstrated the proposed technological tool. Six studies focus on the elicitation stage in conjunction with other RE processes, whereas the other half only target requirements elicitation. The use of technological instruments in RE education has become more prevalent after 2018, as shown by half of the research conducted since then. However, the field is still under-investigated, considering the tremendous advancements in technological tools. Whereas RE studies typically lack controlled experiments to validate their approaches~\cite{daun2017common}, the majority of technology-focused research discusses empirical evaluations including students as participants as shown in Table~\ref{tab:related_work}. Two studies make no mention of the study's participants at all, while the number of participants is relatively low in the others.

The majority of earlier studies are devoted to creating assistive software tools to codify, standardize, and validate the requirements outlined during the requirements elicitation process. Garbers \textit{et. al.} propose an interactive editing tool to support the preparation of correct and consistent software requirements specifications (SRS)~\cite{garbers2006light}. In order to facilitate remote student cooperation on the same SRS document, Liang \textit{et. al.} create a web-based wiki platform~\cite{liang2010experiences}. The problem of ensuring the validity of the generated requirements is attempted in~\cite{ogata2012training} and~\cite{kakeshita2015requirement}. Ogata \textit{et. al.} develop an automated prototype generation tool for a model-based requirements analysis using Unified Modelling Language (UML) to allow the instructor to check the correctness of the input-output data of each section of the model~\cite{ogata2012training}. A more sophisticated tool is suggested in~\cite{kakeshita2015requirement}, which can guide students on the accuracy of the created requirements analysis models by comparing them to the correct one provided by the instructor. A digital game is proposed in~\cite{vega2009training} to teach the steps of elicitation through the metaphoric representation of a software development team meeting. Likewise, educational games have been used in order to make paper-based methods more efficient and engaging while still allowing students to learn real-world concepts~\cite{hainey2011evaluation,yasin2018design}.

The need for a simulated stakeholder in requirements elicitation training is highlighted in a pioneering study presented in~\cite{nakamura2014requirements}. They develop a domain expert system that is capable of monitoring the chat messages of the students working on requirements elicitation of the given project, intervening as needed, inviting students to ask questions, and providing answers by doing keyword-based searches on its domain knowledge. However, the effectiveness of their system was not demonstrated enough due to the toy scenarios employed in the user study, which did not require the students to ask questions to the domain expert system. A serious game is created by \textit{et. al.} to improve understanding and application of the key procedures of requirements engineering methodologies~\cite{garcia2019experiences,garcia2020serious}. The students were provided pre-designed templates of requirements and asked to elicit from and improve upon them. As the elicitation method, interviewing is employed in the game, where students must choose from a pre-loaded set of questions and answers. A similar three-dimensional (3D) serious game was developed in~\cite{ibrahim2019design}, where the game's design decisions are built based on learning objectives of the requirements elicitation process. 

Some recent research focuses on using cutting-edge technologies like virtual reality (VR) or artificial intelligence (AI) based systems to create REET tools. A conversational agent prototype was created by Paschoal \textit{et. al.} to assist students in refining their requirements elicitation skills~\cite{paschoal2018chatterbot}. Their chatbot was designed to be sensitive to the expertise level of students, which can adjust the complexity of the answers accordingly. However, an inconsequential data sample was used in their evaluation study to demonstrate the efficacy of the system. Laiq~\textit{et. al.} proposed a chat-bot based interview simulator built on cloud AI systems that allows the user to engage in natural conversations~\cite{laiq2020chatbot}. Nonetheless, they did not provide a complete experimental study to go into further detail about the system's operation and effectiveness as an interview simulator. In \cite{ochoa2019incorporating}, virtual reality technology is used to create a teaching tool that allows UML modeling in a 3D environment. Likewise, the absence of comprehensive user research prevents us from evaluating the system's suitability for REET. Debnath and Spoletini propose a multi-modal virtual conversational agent playing the role of the stakeholder in order to provide students with an interview simulator that they can use in their free time~\cite{debnath2020designing}. Their proposal also employs a report generator that evaluates user responses and provides a report. Even though their approach is one of the most prominent studies in the literature regarding to addressing the real requirements elicitation interviews' characteristics, the implemented prototype is a quite simplified version of their proposal in terms of interaction capabilities and dialogue generation. Their initial exploratory study demonstrates that participants who used the system before conducting stakeholder interviews outperformed those who did not.

\subsection{Social Robots for Education}
\label{sec:background_int_robot}

Social robots are created to engage with people naturally and personally and to assist them through social interaction. In the field of education, according to the teaching setting and target student group, they are being proposed for use in a variety of roles, including teaching assistant, one-on-one tutor, and peer~\cite{guggemos2022social}. The majority of research concentrate on the early stages of education for specific subjects, including language learning and handwriting~\cite{johal2020research}. This is because employing a robotic system to carry out educational tasks for that level is relatively more straightforward, and younger children often find robots more attractive. Adult learners have higher expectations for the robot's social and technical capabilities to truly feel assisted in the learning process of complicated learning materials~\cite{belpaeme2021social}. Nevertheless, more recent research is focusing on adult learning supported by social robots. Social robots have been demonstrated to bring huge potential to the field by combining technology supported education with enhanced pedagogical approaches like adaptive interactions to address personal needs~\cite{donnermann2022social}, motivational behaviors to engage the students more in the learning activity~\cite{deublein2018scaffolding}, or enhanced social skills to increase attention and fun~\cite{rosenberg2020robot}.   

When looking at the literature on social robot utilization in higher education, there are very few studies that, like ours, conduct user experiment and assess the research objective in a university course setting. Instead, the majority of them were carried out as independent extracurricular activities with university students. Pfeifer and Lugrin investigated the effect of robot gender on learning success in a setup where the robot teaches HTML basics to university students~\cite{pfeifer2018female}. They claimed that the gender of the robot affected the learning success of female students since they learned more when working with a female robot. Rosenberg \textit{et. al.} used a robot as a facilitator in a collaborative group activity, which aims to develop a prototypical application as the human-computer interaction course project. Qualitative analysis results showed that the usage of robots compared to the usage of paper and tablet had been found promising in terms of time management and efficiency. In the comparative study of Velentza \textit{et. al.}, students expressed greater levels of joy and surprise toward the robot instructor than toward the human instructor, but they gained more knowledge in the provided lecture from the latter~\cite{velentza2021learn}. In another study~\cite{velentza2021one}, they looked at how the robot's personality traits and attitudes could affect university students' preferences for further collaborations. The participants, which were involved in a storytelling activity as the experimental setup, favored the robot designed with a positive personality. In~\cite{lei2021effect}, Lei and Rau looked at how participants in a dance-based exercise reacted to either positive or negative feedback on their performances provided by the robot. The findings indicated that the participants appreciated positive feedback more, which might help students feel more motivated, confident, and content. The closest study to our research in terms of being conducted in a field setting of the university context, the authors in \cite{donnermann2022social} conduct a multi-session user study in which students practice several exercises prepared for an undergraduate course. The exercises were prepared in different forms, including single choice, multiple choice, and close texts, and were presented on the tablet interface of the robot. The robot assists the exercise by providing motivational feedback if the student performs correctly. Otherwise, it explains why the user is wrong and gives hints to avoid such mistakes in the future. Focusing more on the effect of adaptability, they also created an adaptive version of the robotic tutor, which personalizes the feedback across multiple sessions according to the user's previous performance. However, in the experimental setting, the robotic tutor's increased adaptability did not appear to improve learning.

\section{\sys{} System}
\label{sec:system}

\begin{figure}[htbp]
\centering
\includegraphics[width=0.6\textwidth]{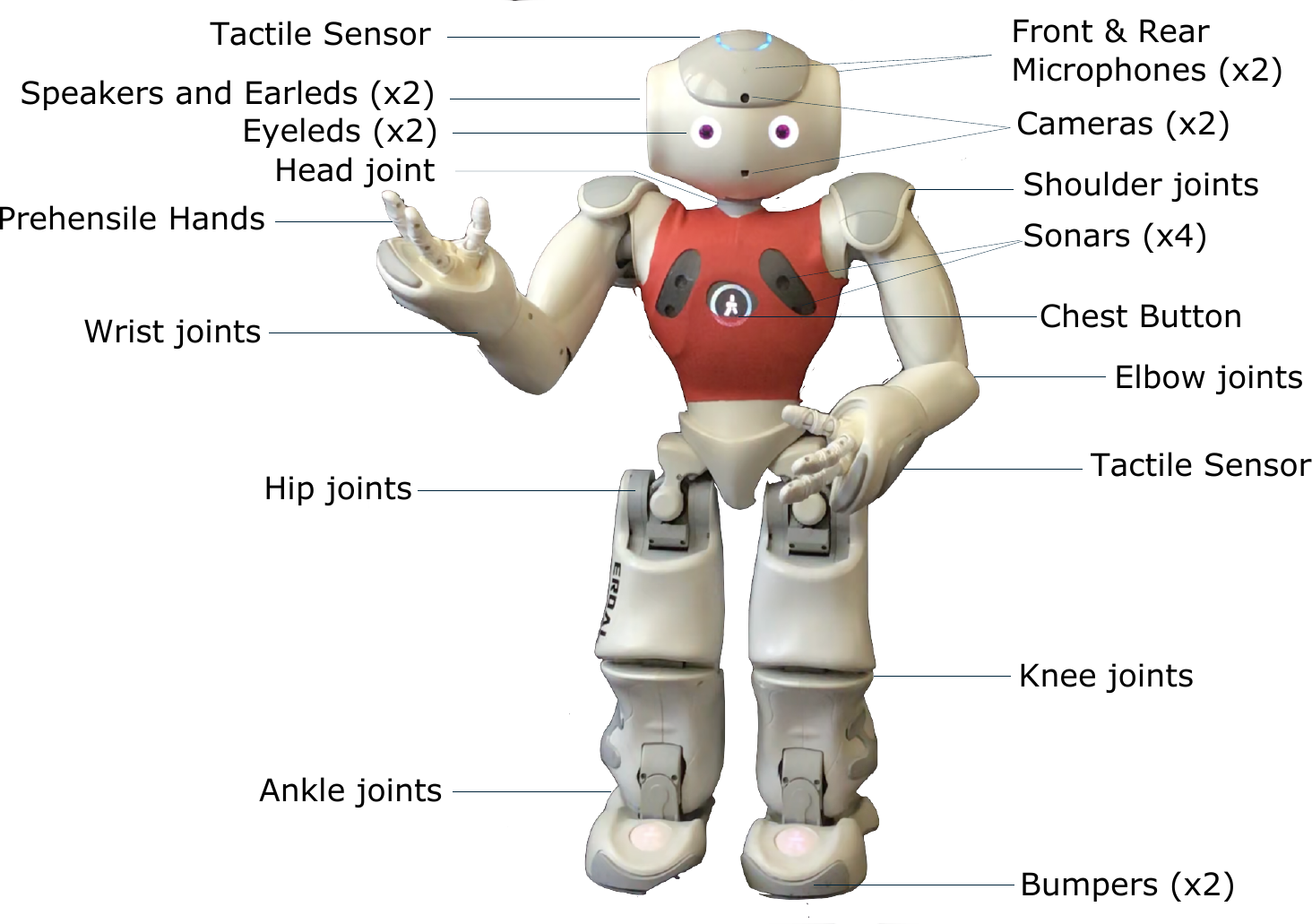}
\caption{The Nao robot.}
\label{fig:nao}
\end{figure}

\sys{} is implemented as a multimodal interactive robotic system that can function fully automatized way. Its system design is grounded in the process of requirements elicitation interview training activities which employ interview practicing with an agent who plays the role of stakeholder and interview performance evaluation by reinforcing feedback on the problematic parts. The system is applied with the Nao robot, which is depicted in Figure~\ref{fig:nao}. Nao is the most widely used humanoid bipedal robot for academic research as it is affordable, easily programmable, and performant~\cite{gouaillier2008nao}. It can be used for social robot-human interaction studies thanks to its attractive design and extensive sensing and acting capabilities. It stands 58 cm tall, is mobile, has auditory and visual perception, and can communicate with people. \sys{}' modular design\footnote{The code repository for the implementation is made available in~\cite{binnur_gorer_2022_7263541}.}, however, allows for the substitution of any of Nao's parts with those of another physical or virtual agent that can perform equivalent tasks, if desired. In this section, we outline the interaction flow of \sys{} along with the functionality of each step. Then, its system architecture is presented.

\subsection{Interaction Flow}
\label{sec:intflow}
This section describes how interviewers in training interact with \sys{}. Each session has four parts. First, \sys{} greets the interviewer and introduces the system. The second part is the interview process in which the robot Nao of \sys{} acts as a stakeholder for the project, following a predefined scenario. At each turn of the interview, \sys{} provides multiple choices for the question of the interviewer's next turn. The scenario unfolds based on the responses of the interviewer. After the interview process, \sys{} proceeds with the feedback process, pointing out the incorrect responses of the interviewer and the rationale. The interviewer is given a second chance to correct the answer. Finally, \sys{} wraps up and ends the training session after the interviewer reviews all the incorrect responses. Figure~\ref{fig:flow} describes the flow of interaction where the interview and feedback processes are explicitly marked. The interaction steps are as follows:

\begin{figure}[htbp]
    \centering
    \includegraphics[width=0.99\textwidth]{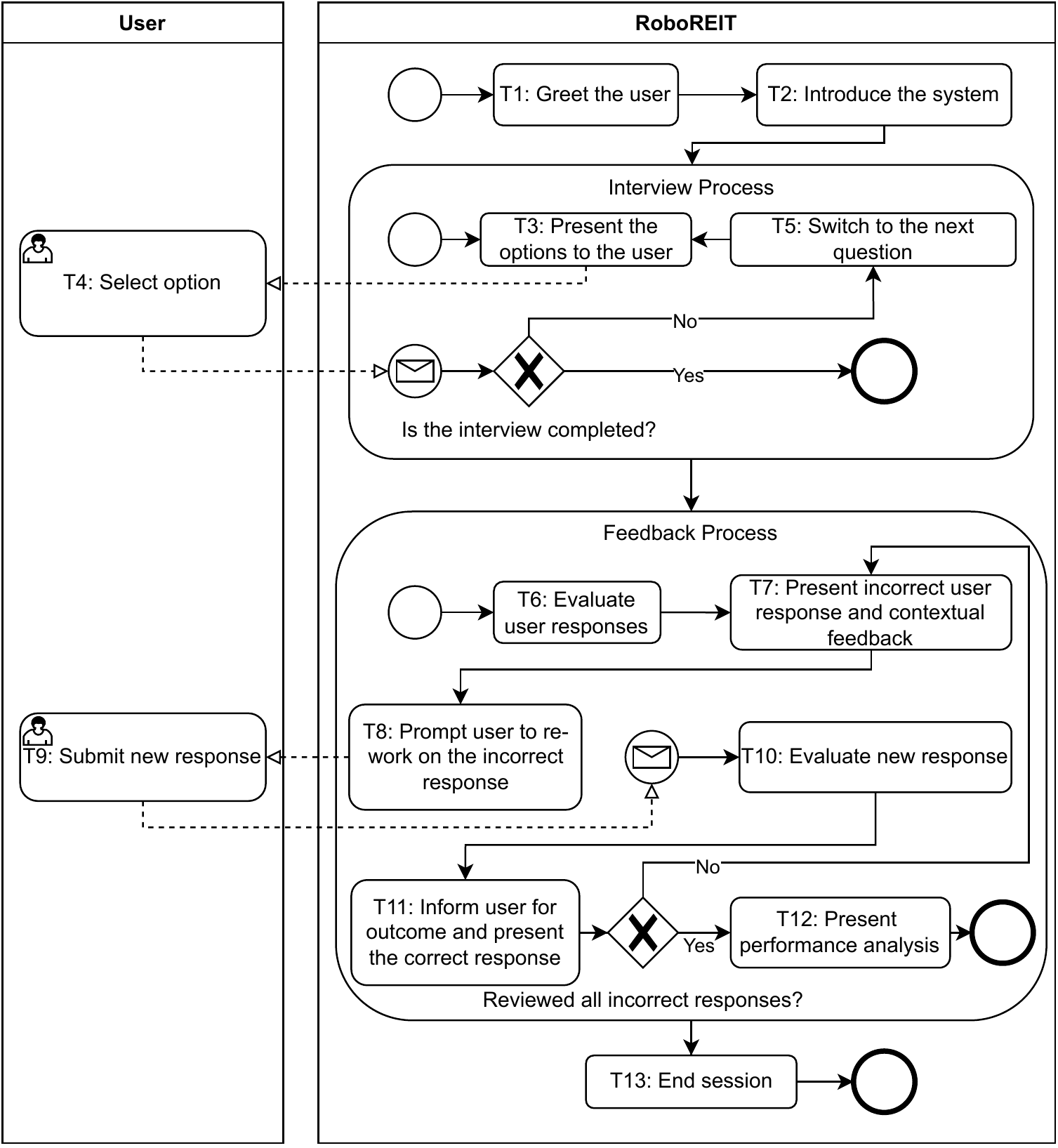}
    \caption{The flow of interaction between the user and \sys{}.}
    \label{fig:flow}
\end{figure}

\begin{description}
    \item[T1: Greet the user.] Nao greets the user to build a rapport so that \sys{} is accepted by the user.
    \item[T2: Introduce the system.] Nao introduces how the system works, sets the role of the user as the interviewer and itself as the interviewee. A brief introduction of the scenario for the session is also provided at this step. 
    \item[T3: Present the options to the user.] At this step, \sys{} displays a screen with potential questions for the interviewer to choose from as presented in Figure~\ref{fig:optScreen_a}. The user is asked to choose one of these questions as the following questions of the interview. This style of interviewer training is also used by Debnath and Spoletini~\cite{debnath2020designing}, where the trainee interacts with a text-based user interface where the response of the stakeholder is also provided as text. In \sys{}, Nao speaks out the response, and only the choices are displayed on a screen. 
    \begin{figure*}
    \centering
        \begin{subfigure}[t]{0.49\textwidth}
            \centering
            \includegraphics[height=1.2in]{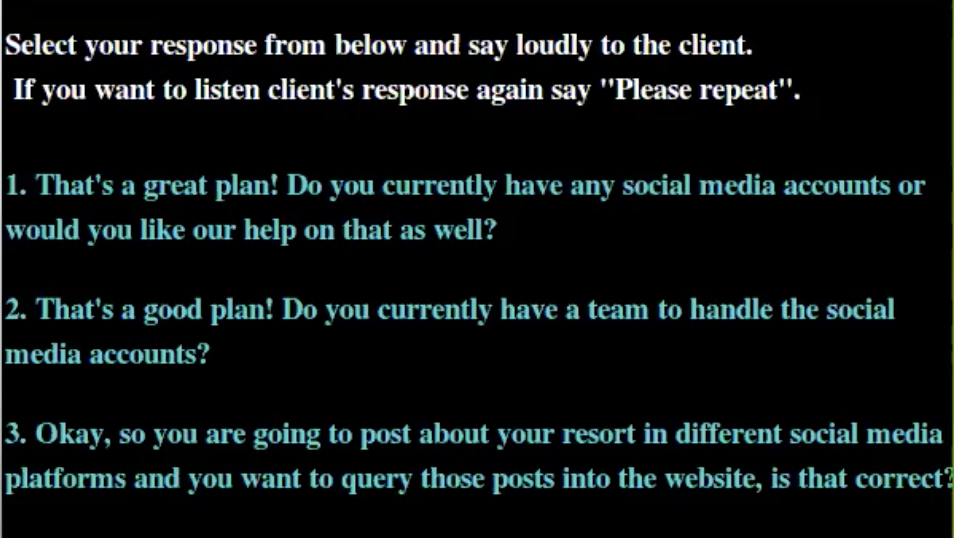}
            \caption{Presenting the user's dialogue options.}
            \label{fig:optScreen_a}
        \end{subfigure}%
        ~ 
        \begin{subfigure}[t]{0.49\textwidth}
            \centering
            \includegraphics[height=1.2in]{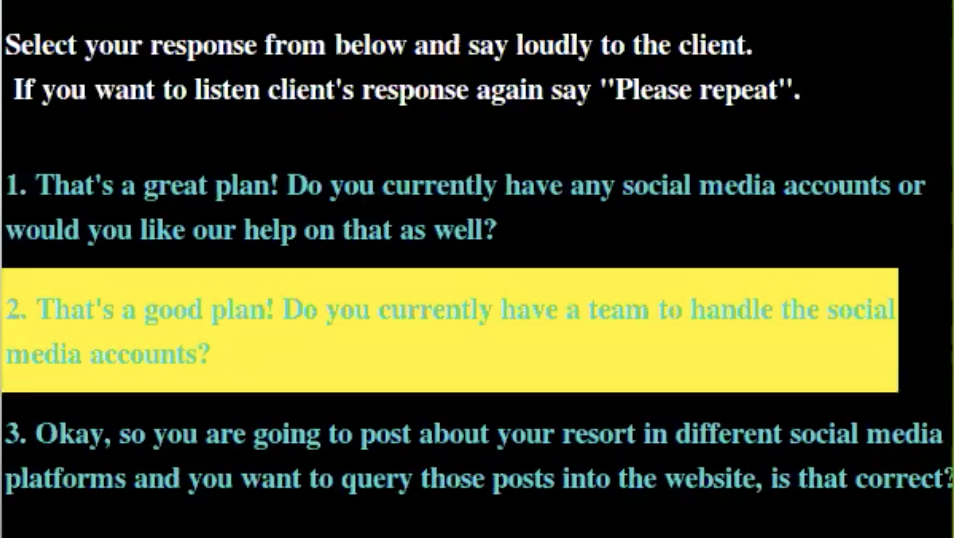}
            \caption{The user choice on the options.}
            \label{fig:optScreen_b}
        \end{subfigure}
    \caption{Sample states of dialogue displayer of \sys{} during the interview session.}
    \label{fig:optScreen}
    \end{figure*}
    \item[T4: Select option.] The user, who is acting as the interviewer, evaluates the interview context considering the response of the stakeholder (Nao) and the direction of the interview. The user then speaks out the option they choose. The chosen option is highlighted on the display to notify the user as shown in Figure~\ref{fig:optScreen_b}. 
    \item[T5: Switch to the next question.] \sys{} recognizes the input from the user and saves the user choice. Nao responds to the question selected by the user. This loop continues as long as there are more questions in the scenario.
     \item[T6: Evaluate user responses.] For this task, \sys{} compares the user responses with the correct options in the predefined scenario and identifies the incorrect responses. Nao informs the user that the feedback session has started. 
     \item[T7: Present incorrect user response and contextual feedback.] \sys{} pre-\\sents the next incorrect response. Nao provides the contextual feedback that is associated with the response's mistake.
     \item[T8: Prompt user to re-work the incorrect response.] Upon providing the feedback, \sys{} gives another chance to the user to identify the correct response along the scenario as shown in Figure~\ref{fig:secondchance_a}.
     \begin{figure*}[htbp]
    \centering
    \begin{subfigure}[t]{0.49\textwidth}
        \centering
        \includegraphics[height=1.35in]{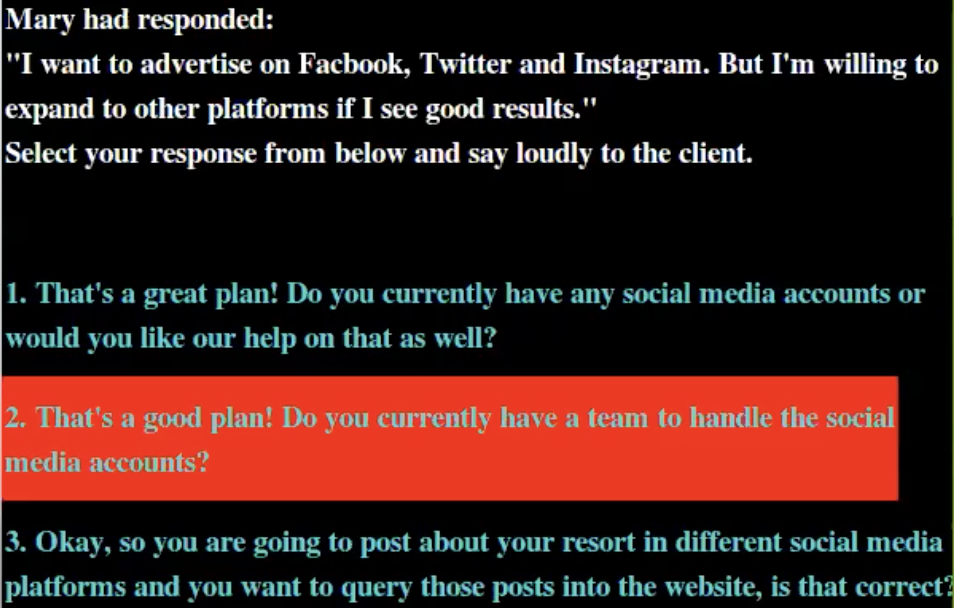}
        \caption{Highlighting the incorrect response of the user and giving a chance to correct it.}
        \label{fig:secondchance_a}
    \end{subfigure}%
    ~
    \begin{subfigure}[t]{0.49\textwidth}
        \centering
        \includegraphics[height=1.35in]{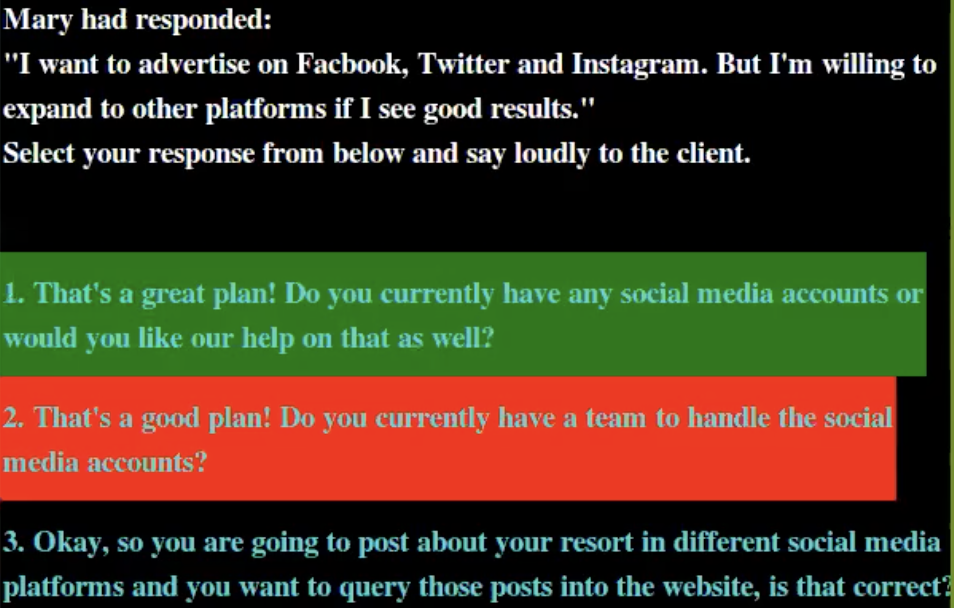}
        \caption{Presenting the evaluation result of the second attempt of the user.}
        \label{fig:secondchance_b}
    \end{subfigure}
    \caption{Sample states of dialogue displayer of \sys{} during the feedback session.}
    \label{fig:secondchance}
    \end{figure*}
     \item[T9:  Submit new response.] The user re-evaluates their options and makes a new choice.
     \item[T10: Evaluate new response.] \sys{} checks whether the new response is correct.
     \item[T11: Inform user of the outcome and present the correct response.] If the new response is correct, \sys{} informs the user as shown in Figure~\ref{fig:secondchance_b}. If the second trial of the user is also wrong, \sys{} highlights the correct option. For the second evaluation, no in-depth feedback similar to step T7 is given to keep the session length within bounds.
     \item[T12: Present performance analysis.] Upon revisiting all incorrect interview questions, Nao notifies the user that the feedback session has ended. \sys{} displays both the number of original incorrect answers and their categories as defined by Bano \emph{et al.}~\cite{bano2019teaching}. The incorrect answers that are fixed after receiving feedback are highlighted as presented in Figure~\ref{fig:overall_feedback}.
     \begin{figure}
         \centering
         \includegraphics[scale=0.4]{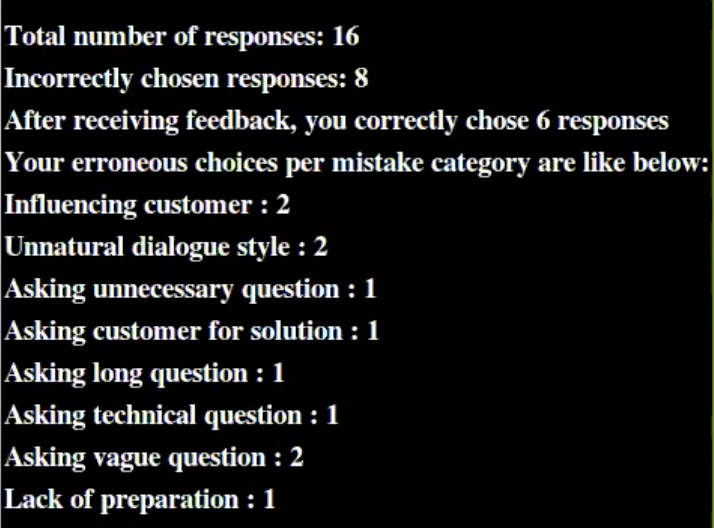}
         \caption{\sys{}'s screen in displaying the overall performance of the user at the end of the session.}
         \label{fig:overall_feedback}
     \end{figure}
     \item[T13: End Session.] Nao thanks the user for participating in the training session and concludes the session. 
\end{description}

\subsection{System Architecture}
\label{sec:modarch}
\begin{figure}[htbp]
\centering
\includegraphics[width=0.8\textwidth]{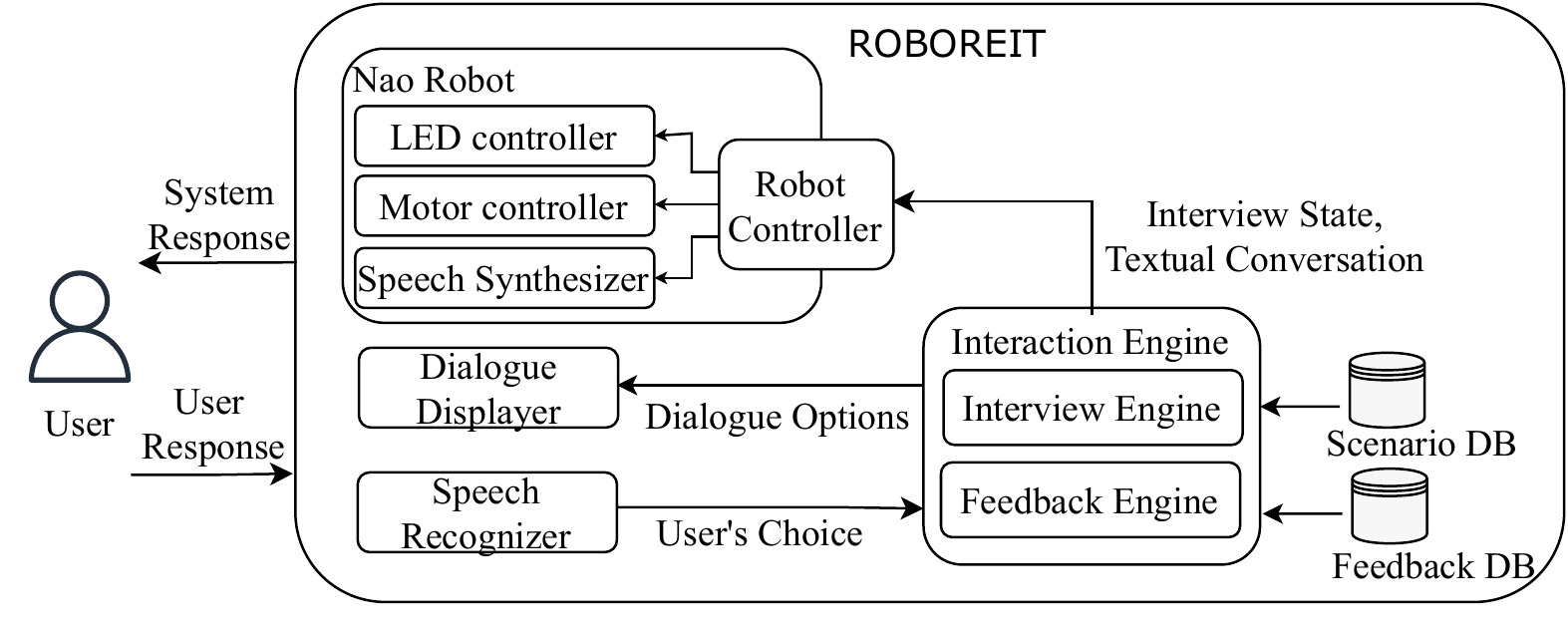}
\caption{The system architecture of \sys{}.}
\label{fig:sys_arch}
\end{figure}

Our system is comprised of four independent modules as shown in Figure~\ref{fig:sys_arch}, namely \textit{Speech Recognizer, Interaction Engine, Dialogue Displayer,} and \textit{Robot Controller}. The first three modules are executed on a standard consumer laptop with modest memory and computing power. \textit{Robot controller} serves as a message interface between the Nao robot and laptop, and is therefore executed on both devices. Except for the speech recognizer module, the entire system is designed to operate on its own.

\paragraph{Speech recognizer} is for processing the speech input into text format and is used to capture the user's input throughout the interview and feedback sessions. In our study, this module was operated in a Wizard of Oz fashion, where a human controller was included in the loop to translate the user's unconstrained speech input (\textit{User Response}) into the expected system input (\textit{User's Choice}).

\paragraph{Interaction engine} manages the interaction flow by iterating over the interaction steps, which are described in Section~\ref{sec:intflow}. The overall interaction is modeled as a finite state machine. The engine decides on the next state according to the current state of the interaction and user input. It contains two internal sub-engines: one for managing the interview and the other for managing the feedback session.
\begin{itemize}
    \item \textit{Interview Engine} queries the scenario database with the given user's response in order to fetch the stakeholder response text. The retrieved text is then sent to the robot controller to be synthesized and spoken to the user. In the meantime, the following question is also collected and passed to the dialogue displayer, where the user will be shown the possible dialogue options as soon as the robot has finished speaking the stakeholder response. 
    \item \textit{Feedback Engine} evaluates each of the interview turns. If the user's choice is incorrect, that interview turn is saved with the preceding stakeholder text, user dialogue options, and the selected incorrect option. The mistake in the incorrect option and the corresponding feedback text for that mistake type are retrieved from the scenario and feedback databases, respectively. In the feedback process, the engine iterates over each saved incorrect interview turn. The preceding stakeholder text and dialogue options, along with the previously selected option are transmitted to the dialogue displayer to remind the user the interview stage. The contextual feedback text is delivered to the robot controller to be uttered by the robot. After the user's second attempt, the feedback engine reevaluates the user's choice. If the second attempt is also inaccurate, no additional feedback is given now. The evaluation result, indicating whether the user's second attempt is correct or incorrect, is communicated to the robot controller and the dialogue displayer to notify the user both verbally and visually.
\end{itemize} 

\paragraph{Dialogue displayer} is a graphical user interface implemented in Python, which shows the dialogue options to the user to select one from at every interview turn (see Figure~\ref{fig:optScreen_a}). The selected option is highlighted with a yellow background when the user makes a choice to let them know that the robot recognized the user's speech (see Figure~\ref{fig:optScreen_b}). The same tool is used in the feedback session to display incorrect interview turns. The stakeholder's previous response and corresponding available options are displayed with the incorrectly chosen option highlighted in red color for every mistaken interview turn (see Figure~\ref{fig:secondchance_a}). After the user's second attempt, the user's choice is highlighted with yellow to notify the user again about the system's recognition of their input. If the selected option is correct this time, the option's yellow background will turn green (see Figure~\ref{fig:secondchance_b}). Otherwise, it is turned into red and the correct option is highlighted with green in order to inform the user. At the end of the overall session, the user is shown the total number of turns made throughout the interview session, the number of turns made erroneously and their categories, and the number of turns that were fixed on the second try (see Figure~\ref{fig:overall_feedback}). In this way, the user can identify the mistake categories with which they struggle and track their development after receiving feedback.

\paragraph{Robot controller} is implemented as a custom engine that controls the robotic platform to realize the requested actions according to the interview state provided by the interview engine. ROS (Robot Operating System) is used to manage the communication and synchronization between the modules which run on the robot and the laptop. In our system, we used Nao as the robotic platform. Nao has two video cameras in the forehead and mouth, providing images resolution up to 1280x960 at 30 frames per second. We implemented a face tracker application that processes the camera images to detect the face of the user and adjust head joint angles accordingly. We also generated accompanying arm and head gestures driven by the motor controller module during the robot's speech to increase liveliness and sociability. Likewise,
eye blinking is implemented using the eye LEDs of Nao to enrich its perceived communicative skill. For the speech synthesizer, we used the built-in module of Nao, which converts a given text to speech and plays it on the robot with customized voice control commands like pitch and speed. All robot controller modules are run on top of Naoqi, the main program of Nao that offers a low-level programming interface that is used to control the robot's joints, adjust LEDs color and intensity, and make the robot speak with the given text in the selected voice. In our study, we used the $4^{th}$ version of Nao, which has the embedded operating system OpenNAO (embedded GNU/Linux based on Gentoo) and middleware NAOqi 2.1.4.  


\section{Evaluation}\label{sec:evaluation}

This section presents the design, the execution procedure, and the empirical evaluation results of the user study of \sys{}. Our goal is to explore \sys{} for its improved design in comparison to a simpler system by collecting users' opinions on the system's utility and acceptability in requirements elicitation interview training. We use \vico{}, provided in \cite{debnath2020designing}, as the reference system. We would like to perform an exploratory analysis to assess user perceptions and experiences for the two systems. To this end, we designed a user study with two experimental groups. The first group (Group A) practiced a requirements elicitation interview with \sys{} and trained throughout the feedback session by rehearsing the faulty turns of the interview, while the second group (Group B) used \vico{} to conduct an interview and received a mistake report at the end. We addressed the following research questions (\textbf{RQs}) in our study:

\begin{enumerate}[start=1,label={\bfseries RQ\arabic*:}, leftmargin = 3em]
    \item \textit{How does the form of the interaction (audio-visual interaction with a physical robot in case of \sys{} and mouse click-based interaction with a web agent in case of \vico{}), which the user is exposed to while conducting the interview, influence the speed of user response?} \\
    Constant practicing in requirements elicitation interview training is highly important. Although theoretical knowledge, standard procedures, and guidelines are taught to the students in the lecture, the lack of a combination of these with real-world practice prevents the students from excelling in conducting successful interviews. It is essential to use the natural interaction style, such as in human-human communication, to build more realistic interview systems. Interview flow management, as one of the important soft skills in conducting requirements elicitation interviews, requires maintaining communication without compromising stakeholder engagement. So we analyzed how the form of the interaction and interacted entity stimulate the user to keep a smoother communication flow by measuring the average processing speed of the users throughout the interview.  
    
    \item \textit{How do \vico{} and \sys{} influence the perceived acceptance of the underlying technology in the dimensions of \textbf{RQ2a:} perceived attitudes, \textbf{RQ2b:} perceived ease-of-use, \textbf{RQ2c:} perceived usefulness}?\\
    This RQ aims to assess the users' acceptance model of the underlying technology to be used for requirements elicitation interview training. The participants are asked to evaluate perceived attitudes, perceived ease-of-use, and perceived usefulness using a 5-point Likert scale via an extended version of the technology acceptance model survey explained in Section \ref{sec:materials}. Hence, all variables used to answer this RQ take integer values in $\{1, ... ,5\}$, where higher values indicate higher positive attitudes, ease-of-use, and usefulness. 
    
    \item \textit{How do \vico{} and \sys{} influence the perceived engagement of training with the system?} \\
    A tutoring system's ability to maintain the user's connectivity and interest high is crucial for maximizing the learning outcome. Engagement typically has multiple components in the context of education, like behavioral, cognitive, and affective engagement \cite{appleton2006measuring}. In our study, we focus on behavioral engagement which stands for positive conduct to the education activity. To answer this RQ, we collected user's opinion on a 5-point Likert scale on how much they feel engaged while using the system.  
    \item \textit{How do the feedback components of \vico{} and \sys{} influence the perceived effectiveness and helpfulness of the system to identify the mistakes?} \\
    The RQ aims to understand whether users consider the feedback component helpful. The participants were asked two questions on a 5-point Likert scale in the post experiments questionnaire to rate the effectiveness of the feedback component and the helpfulness in identifying the mistakes.
\end{enumerate}

\subsection{Study Design}\label{sec:study_design}
\begin{figure*}[htbp]
   \centering
	\includegraphics[scale=0.35]{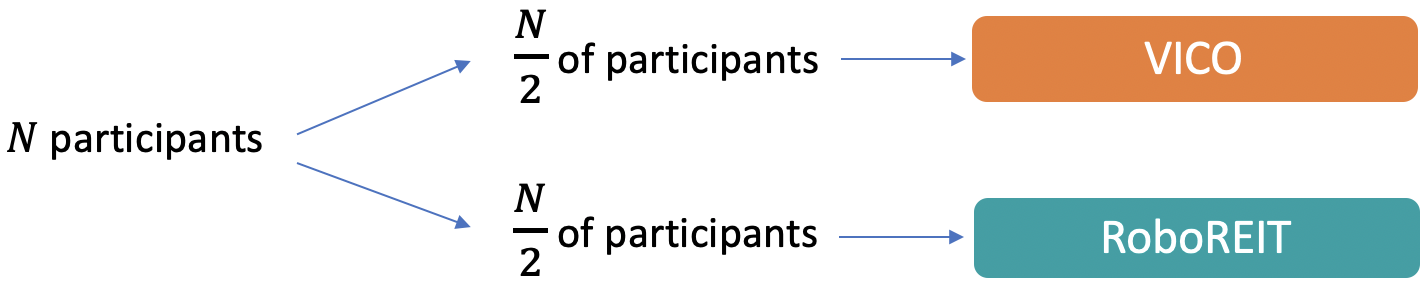}
	\caption{The between-subjects experimental design of the user study.}
	\label{fig:study_design}
\end{figure*}

To investigate the user experience of \sys{}, we designed a between-participants user study by following the guidelines in Hoffman \textit{et al.}~\cite{hoffman2020primer}. The between-subject condition is the system itself, the \vico{} or the \sys{}. Participants are randomly assigned to one of the two setups, \vico{} or \sys{} as shown in Figure \ref{fig:study_design}. The experiment includes training in requirements elicitation interviews with the given system. The same interview scenario is used for both conditions.

\subsubsection{Procedure}
\label{sec:procedure}

Due to Covid-19 pandemic restrictions, the experiment is planned to be conducted online through a video conference tool. The experiment announcement is prepared as a written document and video to be posted on the social media channel for the RE class as well as in graduate student email groups. The overall study description, the informed consent form, the necessary technical setup, and the anticipated experiment duration are all included in the document so that potential participants can assess their availability. The informed consent form concerns data protection and permission to record on video during the study. The experimenter contacts the candidate to set up an experiment time if the candidate fits the criteria outlined in the description document, approve the consent, and agree to participate. The experiment is prepared in English because it is the language of both education and business. Likewise, all introductory materials, the informed consent form, and the administered questionnaires are in English.

Throughout the entire study, the experimenter is required to stay in the environment to observe the study and handle the speech-to-text processing component in \sys{} condition. Since the target participant population's native language is not English, they are not expected to be highly professional in spoken English. That's why we would like to have the experimenter as a wizard for the speech-to-text component to ensure that the system perceives the correctly processed verbal responses of the participant. However, the participant is not aware of the experimenter's presence and is explicitly informed before the study that the experimenter would be leaving the environment after initiating the study and that no help or assistance would be provided during the experiment. Participants are asked to complete the experiment on their own. The goal is to avoid any participant bias caused by the presence of others. Participant bias can occur in different forms, such as social desirability when participants desire to present themselves in the best look possible and may not be completely honest in their reactions or comments \cite{king2000social}. Likewise, participants who are aware that they are taking part and being observed in a research study could alter their behavior or responses in accordance with what they think the experimenter desires.

The experimental procedure has the following steps:
\begin{enumerate}
    \item At the time of the experiment, the experimenter opens the video conference call. When the participant joins, the experimenter mentions the experiment's flow and answers the participant's questions, if any. Then, a series of pre-experiment questionnaires is shared with the participant to be completed before starting the experiment. After completing the questionnaires, the participant is expected to notify the experimenter that they are ready to start the study.
    \item \begin{itemize}
        \item[a)] \vico{} condition: There are no video or audio interaction modalities in this condition as the participant conducts the interview by clicking on the provided web interface. The participant is provided the link for \vico\footnote{http://www.interviewsim.com.s3-website.us-east-2.amazonaws.com/} and asked to access the link and to share their screen on the video conference tool so that the entire interview flow is visible and recordable. The experimenter then informs the subject that they may start the study and leaves the environment. 
        \item[b)] \sys{} condition: The participant is made aware of the experiment's physical requirements before the study begins. The experimenter ensures that the face of the participant is visible and they can be heard clearly.
        The dialogue displayer of \sys{}, which includes the live video of the robot and the dialogue options, is then presented on the shared screen by the experimenter. After receiving the participant's confirmation that they are ready to begin, the experimenter starts \sys{} and informs the participant that she will leave the environment. She remains in the experiment environment, though, to control the speech-to-text component of \sys{} but not visible to the participant at all. She provides the selected option ID to the system after each verbal response of the participant.
    \end{itemize}
    \item After the session is completed, the experimenter returns back to the call and requests the participant to fill out the post-experiment questionnaires.
    \item The experiment is concluded once the study and questionnaires are completed. As closing remarks, the experimenter thanks the participant and reacts to any subsequent comments or inquiries from the participant.
\end{enumerate}

We conducted three pilot trials with our colleagues to evaluate the system and more accurately predict the timings of the experimental stages. The overall experiment is expected to take approximately 25 minutes and 45 minutes for \vico{} and \sys{} conditions, respectively. The pre-experiment questionnaire takes about 8 minutes to complete, whereas the post-experiment survey takes 5 minutes. However, given that the length of the feedback phase will depend on how many mistakes the participant makes, we anticipate some variations in the overall experiment lengths, particularly for the \sys{} condition.

\begin{figure*}[htbp]
\centering
\includegraphics[scale=0.45]{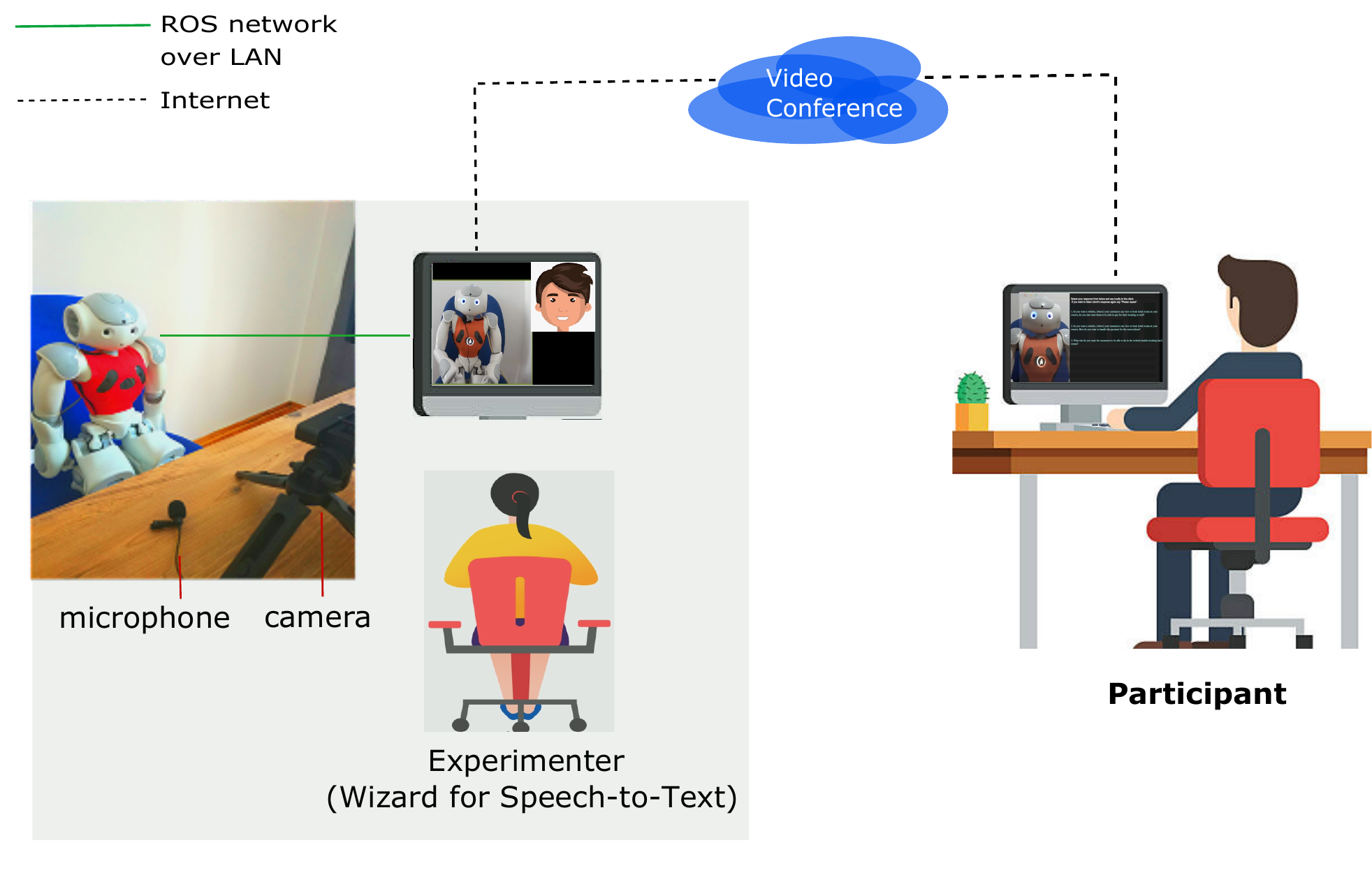}
\caption{Experimental setup for \sys{}.}
\label{fig:setup}
\end{figure*}

The study is set up in a room that is equipped with a camera, a lapel microphone, the Nao robot sitting on a chair, and the laptop that is used to execute the system components (see Section~\ref{sec:modarch}). The experimenter is in the same room to initiate the video call and control the study but is not visible to the participant. The participant is requested to join the experiment on a computer with a stable network connection in a well-lighted and quiet environment to minimize environmental distractions. The topological overview of the overall experimental setup is given in Figure \ref{fig:setup}.

\subsubsection{Materials}
\label{sec:materials}
The materials used in the experimental procedure can be found in our shared repository~\cite{binnur_gorer_2022_7263541}. 
\sectopic{Scenario.} We used the Cool Ski Resort scenario provided in \cite{ferrari2020sapeer}. The scenario context is tailored from a real-world use case, designing a website and its connected technological system for a resort company. The goal is to elicit requirements for the website design, social media connection, and database utilization. 

The interview for this scenario is structured as a dialogue in which the stakeholder and the requirements engineer exchange responses in turn. At each turn, the requirements engineer has three options to select from: two of them are manipulated with some mistakes, and the other one is the correct response. The requirements engineer choice decides the stakeholder response and following requirements engineer options for the next turn. Hence, the overall dialogue flow is shaped by the requirements engineer's responses.
The minimum and maximum turn counts are 15 and 19, respectively. We also checked the induced mistake types for all possible requirements engineer responses. The induced mistake counts per mistake class are shown in Table \ref{tab:scenario_mistakes}.

\begin{table*}
\caption{The induced mistake occurrences in the scenario with their associated mistake class as described in~\cite{bano2019teaching}.}
\label{tab:scenario_mistakes}
\begin{tabular}{lllr}
\hline
\textbf{ID} & \textbf{Mistake Type}             & \textbf{Associated Mistake Class} & \textbf{Occurance} \\ \hline
1           & Lack of preparation               & \textit{Teamwork and Planning}    & 4                  \\
2           & Lack of planning                  & \textit{Teamwork and Planning}    & 3                  \\ \hline
3           & Not identifying stakeholders      & \textit{Question Omission}        & 1                  \\
4           & Not asking about existing system  & \textit{Question Omission}        & 6                  \\ \hline
5           & Asking long question              & \textit{Question Formulation}     & 3                  \\
6           & Asking unnecessary question       & \textit{Question Formulation}     & 7                  \\
7           & Asking stakeholder for solution      & \textit{Question Formulation}     & 15                 \\
8           & Asking vague question             & \textit{Question Formulation}     & 32                 \\
9           & Asking technical question         & \textit{Question Formulation}     & 5                  \\ \hline
10          & Incorrect ending of the interview & \textit{Order of interview}       & 6                  \\ \hline
11          & Influencing stakeholder              & \textit{stakeholder interaction}     & 9                  \\
12          & No rapport with stakeholder          & \textit{stakeholder interaction}     & 16                 \\
13          & Unnatural dialogue style          & \textit{Communication skills}     & 11                 \\ \hline
\end{tabular}
\end{table*}

\sectopic{Pre-experiment survey.} We collect participants' demographics and their opinions about the robots.
\begin{itemize}
    \item \textit{Demographics:} We took a general demographic survey of the participants (e.g., age, gender, current occupation, and years of working experience). We also inquired about their level of training and experience in conducting software requirements elicitation interviews with the questionnaire in Table~\ref{tab:demog_training}.
    \item \textit{Negative Attitudes towards Robots:} We measured the negative attitude towards robots which could have a powerful influence on our results. The ``Negative Attitude toward Robot Scale" (NARS) measures people's anxiety towards robots. We used the version provided in~\cite{syrdal2009negative}.
\end{itemize}

\begin{table*}[htbp]
\caption{Questionnaire for the level of experience and expertise in conducting requirements elicitation interviews.}
\label{tab:demog_training}
\begin{tabular}{|l|lcccc|}
\hline
\rowcolor[HTML]{EFEFEF} 
\textbf{Q1} & \multicolumn{5}{l|}{\cellcolor[HTML]{EFEFEF}\begin{tabular}[c]{@{}l@{}}How many times have you practiced requirements elicitation interview so far?\\ (including those for class projects and other purposes)\end{tabular}}                                                                                                                                                                                                          \\ \hline
              & \multicolumn{1}{c|}{\textit{0}}                                                              & \multicolumn{1}{c|}{\textit{1-3}}                                                          & \multicolumn{1}{c|}{4-6}                                                                           & \multicolumn{2}{c|}{more than 6}                                                                                                                              \\ \hline
\rowcolor[HTML]{EFEFEF} 
\textbf{Q2} & \multicolumn{5}{l|}{\cellcolor[HTML]{EFEFEF}How many real requirements elicitation interviews have you conducted so far?}                                                                                                                                                                                                                                                                                                                                      \\ \hline
              & \multicolumn{1}{c|}{\textit{0}}                                                              & \multicolumn{1}{c|}{\textit{1-3}}                                                          & \multicolumn{1}{c|}{\textit{4-6}}                                                                  & \multicolumn{2}{c|}{\textit{more than 6}}                                                                                                                     \\ \hline
\rowcolor[HTML]{EFEFEF} 
\textbf{Q3} & \multicolumn{5}{l|}{\cellcolor[HTML]{EFEFEF}How did you learn how to conduct requirement elicitation interviews?}                                                                                                                                                                                                                                                                                                                                              \\ \hline
              & \multicolumn{1}{c|}{\textit{Self study}}                                                     & \multicolumn{1}{c|}{\textit{\begin{tabular}[c]{@{}c@{}}University \\ course\end{tabular}}} & \multicolumn{1}{c|}{\textit{\begin{tabular}[c]{@{}c@{}}Company\\ training\end{tabular}}}           & \multicolumn{2}{c|}{\textit{\begin{tabular}[c]{@{}c@{}}I have no \\ training on this\end{tabular}}}                                                           \\ \hline
\rowcolor[HTML]{EFEFEF} 
\textbf{Q4} & \multicolumn{5}{l|}{\cellcolor[HTML]{EFEFEF}I have good theoretical understanding of requirements elicitation interviews.}                                                                                                                                                                                                                                                                                                                                \\ \hline
              & \multicolumn{1}{c|}{\textit{\begin{tabular}[c]{@{}c@{}}strongly\\ disagree(1)\end{tabular}}} & \multicolumn{1}{c|}{\textit{\begin{tabular}[c]{@{}c@{}}slightly\\ disagree\end{tabular}}}  & \multicolumn{1}{c|}{\textit{\begin{tabular}[c]{@{}c@{}}neither agree\\ nor disagree\end{tabular}}} & \multicolumn{1}{c|}{\textit{\begin{tabular}[c]{@{}c@{}}slightly\\ agree\end{tabular}}} & \textit{\begin{tabular}[c]{@{}c@{}}strongly\\ agree(5)\end{tabular}} \\ \hline
\rowcolor[HTML]{EFEFEF} 
\textbf{Q5} & \multicolumn{5}{l|}{\cellcolor[HTML]{EFEFEF}I feel confident and comfortable in practicing requirement elicitation interviews.}                                                                                                                                                                                                                                                                                                                                \\ \hline
              & \multicolumn{1}{c|}{\textit{\begin{tabular}[c]{@{}c@{}}strongly\\ disagree(1)\end{tabular}}} & \multicolumn{1}{c|}{\textit{\begin{tabular}[c]{@{}c@{}}slightly\\ disagree\end{tabular}}}  & \multicolumn{1}{c|}{\textit{\begin{tabular}[c]{@{}c@{}}neither agree\\ nor disagree\end{tabular}}} & \multicolumn{1}{c|}{\textit{\begin{tabular}[c]{@{}c@{}}slightly\\ agree\end{tabular}}} & \textit{\begin{tabular}[c]{@{}c@{}}strongly\\ agree(5)\end{tabular}} \\ \hline
\end{tabular}
\end{table*}

\sectopic{Post-experiment survey.} We collect participants' opinions about the system experience.
\begin{itemize}
\setlength\itemsep{1em}
    \item \textit{Questionnaire on System Design:} We asked participants how engaged they feel while using the interview system. We also looked at how helpful they found the system to identify and learn from the mistakes by asking a) the system's helpfulness in identifying the mistakes b) the feedback component's effectiveness.
    The questionnaire is shown in Table \ref{tab:ques_design}, which is adapted from~\cite{alessio_ferrari_2020_3765214}. 
    \item \textit{Technology Acceptance Model:} We used the expanded version of the technology acceptance model (TAM) proposed in Yang \textit{et. al.}~\cite{yang2004s}. TAM was originally presented by Davis \textit{et. al.} to explain why people adopt information technology to execute tasks. The model revealed two key attitudes that drive information system utilization: perceived usefulness and ease-of-use. Yang \textit{et. al.}~\cite{yang2004s} refined it by also taking into account the affective and cognitive aspects of attitude. 
        \begin{itemize}
        \setlength\itemsep{1em}
            \item[] a) Attitudes towards using: Yang~\textit{et. al.} expand TAM by incorporating attitudes toward using the technology. Their study implies that attitudes influence users' technology acceptance.
            \item[] b) Perceived ease-of-use: refers to ``the degree to which a person believes that using a particular system would be free of effort''. Physical and/or mental effort, as well as an exertion to learn how to operate the technology, might all be included in the effort.
            \item[] c) Perceived usefulness: is defined as ``the degree to which a person believes that using a particular system would enhance their job performance''. It has to do with job effectiveness, productivity (saving time), and the system's relative value to one's job.
        \end{itemize}
    \item \textit{General comments:} We asked participants for their favorite and least favorite aspects of the system and suggestions for how to improve it. The responses are collected as open-ended text and examined qualitatively in Section \ref{sec:conclusion}.
\end{itemize}

\begin{table}[htbp]
\caption{Questionnaire for the system design.}
\label{tab:ques_design}
\footnotesize
\centering
\begin{tabular}{|lcccc|}
\hline
\multicolumn{1}{|c|}{\begin{tabular}[c]{@{}c@{}}Strongly\\ Disagree\\ 1\end{tabular}} & \multicolumn{1}{c|}{\begin{tabular}[c]{@{}c@{}}Slightly \\ Disagree\\ 2\end{tabular}} & \multicolumn{1}{c|}{\begin{tabular}[c]{@{}c@{}}Neither Agree \\ nor Disagree\\ 3\end{tabular}} & \multicolumn{1}{c|}{\begin{tabular}[c]{@{}c@{}}Slightly\\ Agree\\ 4\end{tabular}} & \begin{tabular}[c]{@{}c@{}}Strongly \\ Agree\\ 5\end{tabular} \\ \hline
\multicolumn{5}{|l|}{I was engaged while using the interview system.}                                                                                                                                                                                                                                                                                                                                                              \\ \hline
\multicolumn{5}{|l|}{\begin{tabular}[c]{@{}l@{}}I found the interview system helpful to identify the mistakes \\ in my responses.\end{tabular}}                                                                                                                                                                                                                                                                                    \\ \hline
\multicolumn{5}{|l|}{I found the interview system's dialogue options to be practical.}                                                                                                                                                                                                                                                                                                                                             \\ \hline
\multicolumn{5}{|l|}{I found the interview system's feedback component to be effective.}                                                                                                                                                                                                                                                                                                                                           \\ \hline
\end{tabular}
\end{table}

\subsubsection{Dependent Variables}

The dependent variables arising from the RQs are \textit{processing speed} (RQ1), \textit{perceived acceptance} (RQ2), \textit{perceived engagement} (RQ3), and \textit{perceived helpfulness in spotting mistakes}. Their formal definition is given below.

\paragraph{Processing speed} Cognitive processing speed is a measure of how quickly a person can receive information and react to their environment~\cite{salthouse1996processing}. It is significantly impacted by the cognitive job the person is doing (such as driving, reading, or conducting an interview), the time constraints placed on the task, and the total social interaction the person is managing while performing the work. We designed a measure \textit{processing speed} ($PS$) to quantify the processing speed of the participants in each interview turn. We investigate the discrepancies in the two experimental groups, \vico{} and \sys{}, which employ setups with different levels of social complexity.

Let $P^A$ and $P^B$ be the set of participants in group $A$ and group $B$, respectively. A participant $p \in \{P^A \cup P^B\}$ performs an interview $I(p)$. An interview $I(p)$ consists of $T(p)$ number of turns. Each turn $t(p)$ has $T3$, $T4$, $T5$ steps (see Figure \ref{fig:flow}), with $t(p) \in \{1...|T(p)|\}$. The participant is presented with the available options to select from in step $T3(t(p))$. After evaluating the options, the participant selects one and responds in $T4(t(p))$. Then, the stakeholder responds back in $T5(t(p))$. $PS_{t}$ is determined by dividing the defined cognitive load $CL_{t}(p)$ by the length of time $RT_{t}(p)$ needed for participant $p$ to handle it.
\begin{equation}
PS_{t}(p) = \dfrac{CL_{t}(p)}{RT_{t}(p)}
\end{equation}

We measure $PS$ from the point at which the participant is given the options until they respond. During this interval, the participant is expected to read the presented multiple-choice options and select one of them as the appropriate response. This constitutes the cognitive load $CL_{t}(p)$ in turn $t$ for participant $p$. For the \sys{} condition, response time $RT$ is measured as the interval after the robotic stakeholder speaks its speech, and the options are presented to the participant (step $T3$) until the participant provides the response (step $T4$). In the \vico{} condition, response time is measured as the interval after the simulator web page is refreshed with the stakeholder speech and user options until the participant clicks on one of the options. That is why $CL$ in \vico{} condition also includes the load brought by reading the stakeholder speech text,

\begin{equation}
CL_{t(p)} =
  \begin{cases}
    eval(opt_{t(p)})  & \quad \text{if } p \in \text{group A} \\[0.5ex]
    eval(opt_{t(p)}) + eval(cliResp_{t(p)})       &
    \quad \text{if } p \in \text{group B}
  \end{cases}
\end{equation}
which is quantified as the number of words in the text.
\begin{equation}
eval(cli\_resp_{t(p)}) = length(cli\_resp_{t(p)})
\end{equation}

We base on the design definitions of multiple choice questions in the education literature to quantify the evaluation of options. The structure of multiple choice questions is composed of the difficulty index, discrimination index, and distractor efficiency \cite{MOUSSA1991283}. The difficulty index indicates how tough it is to choose the right answer from the options. One of the aspects of the difficulty index is the similarity of the options ($S$) \cite{ascalon2007distractor,shin2019multiple}, which makes the elimination harder and requires the user to read the options multiple times to identify the differences. Based on that, we measure the cognitive load required to evaluate the options as the total amount of words in the options weighted with their similarity score.
\begin{equation}
    eval(opt_{t(p)}) = (1 + S(opt_{t(p)})) * length(opt_{t(p)})
\end{equation}
We used universal sentence encoder \cite{cer2018universal} in order to compute the pairwise semantic similarity between the items in a given option set. Universal sentence encoder is one of the most accurate tools to find the similarity between two pieces of text. It provides a pre-trained model\footnote{https://tfhub.dev/google/universal-sentence-encoder/4} that is ready to use in an application without a need to train from scratch. The tool converts any provided text to a fixed-length vector representation. Then, we calculate cosine similarity between the two vectors of $opt\_i$ and $opt\_j$ to get a similarity score $sim(opt\_i, opt\_j)$ in $[0, 1]$, where $i \ne j$ and $i,j \in \{1, 2, 3\}$ as we provide three options in each turn. The maximum of the three pair-wise similarities is used as the similarity level $S(opt)$ for that option set. 
\begin{equation}
S(opt_{t(p)}) = \max_{ (i, j) \in \{1, 2, 3\}, i \ne j} sim(opt\_i, opt\_j) 
\end{equation}
The average processing speed for the interview of participant $p$ is then given by averaging $PS_t(p)$ over all the turns:
\begin{equation}
PS(I(p)) = \dfrac{1}{T(p)} \sum_{t \in \{1...|T(p)|\}} {PS_t(p)}
\end{equation}

\paragraph{Perceived acceptance} We investigated the perceptions of the participant $p \in P$ for the acceptance of the given technology. We measured the perceived acceptance by three variables; attitudes $ATT(p)$, perceived ease-of-use $PEU(p)$, and perceived usefulness $PU(p)$. The participants' scores are collected on the technology acceptance model questionnaire upon the completion of the training session. All the variables take values in {1,...,5}, where higher values indicate higher acceptance.

\paragraph{Perceived engagement} The engagement score $PE(p)$ is given by participant $p$ by the post-experiment questionnaire prepared for the system design evaluation. The variable $e$ takes values in {1,...,5}, where higher values indicate higher engagement.

\paragraph{Perceived helpfulness in spotting mistakes} As for the variable helpfulness of the system in spotting mistakes $PH(p)$, we collected participant scores on this by the post-experiment questionnaire. It takes values in {1,...,5}, where higher values indicate higher helpfulness in spotting mistakes.

\subsection{Study Execution}
\label{sec:study_execution}
In order to attract people to take part in the experiment, we created a short animated video that describes the study's overall goal, who is eligible to participate, how it will be carried out, and how long it will take. The video did not include any information about either the manipulation effect of the experiment or the scenario details used in the interview to avoid any bias. We shared the video with the relative groups and people who, we think, might be interested in the experiment. Twenty-four participants, nine women and fifteen men, responded positively.

\renewcommand{\arraystretch}{1.2}
\begin{table*}
\scriptsize
\centering
\caption{Demographics data of the participants for the two experimental groups.}
\label{tab:participant_demographics}
\begin{tabular}{llcc}
\hline
                                                                                                                                   & \textit{}                                           & \textbf{\begin{tabular}[c]{@{}c@{}}\vico{}    \\ Group\end{tabular}} & \textbf{\begin{tabular}[c]{@{}c@{}}\sys{} \\ Group\end{tabular}} \\ \hline
Participants                                                                                                                       & \multicolumn{1}{l|}{\textit{count}}                 & 12                                                                     & 12                                                                       \\ \hline
\multirow{2}{*}{Age(years)}                                                                                                        & \multicolumn{1}{l|}{\textit{range}}                 & 25 - 41                                                                & 18 - 37                                                                  \\
                                                                                                                                   & \multicolumn{1}{l|}{\textit{M (SD)}}             & 30 (4.83)                                                              & 29.2 (5.45)                                                              \\ \hline
\multirow{2}{*}{Gender}                                                                                                            & \multicolumn{1}{l|}{\textit{Female}}                & 5                                                                      & 4                                                                        \\
                                                                                                                                   & \multicolumn{1}{l|}{\textit{Male}}                  & 7                                                                      & 8                                                                        \\ \hline
\multirow{3}{*}{Occupation}                                                                                                        & \multicolumn{1}{l|}{\textit{IT professsional}}      & 2                                                                      & 2                                                                        \\
                                                                                                                                   & \multicolumn{1}{l|}{\textit{Grad Student}}          & 2                                                                      & 0                                                                        \\
                                                                                                                                   & \multicolumn{1}{l|}{\textit{Both}}                  & 8                                                                      & 10                                                                       \\ \hline
\multirow{4}{*}{\begin{tabular}[c]{@{}l@{}}Years of Work \\ Experience\end{tabular}}                                               & \multicolumn{1}{l|}{\textit{0 years:}}              & 3                                                                      & 0                                                                        \\
                                                                                                                                   & \multicolumn{1}{l|}{\textit{1-3 years:}}            & 4                                                                      & 2                                                                        \\
                                                                                                                                   & \multicolumn{1}{l|}{\textit{4-6 years:}}            & 1                                                                      & 6                                                                        \\
                                                                                                                                   & \multicolumn{1}{l|}{\textit{\textgreater 6 years:}} & 4                                                                      & 4                                                                        \\ \hline
\begin{tabular}[c]{@{}l@{}}Self reported confidence level\\ in practicing RE interviews\\ (1-lowest, 5-highest)\end{tabular}       & \multicolumn{1}{l|}{\textit{M (SD)}}             & 3 (1.12)                                                               & 3.25 (0.62)                                                              \\ \hline
\begin{tabular}[c]{@{}l@{}}Self reported level of theoretical   \\ knowledge on RE interviews\\ (1-lowest, 5-highest)\end{tabular} & \multicolumn{1}{l|}{\textit{M (SD)}}             & 2.58 (0.99)                                                            & 3.41 (0.66)                                                              \\ \hline
\multirow{4}{*}{\begin{tabular}[c]{@{}l@{}}\# of practiced \\ RE interviews\end{tabular}}                                          & \multicolumn{1}{l|}{\textit{0}}                     & 4                                                                      & 2                                                                        \\
                                                                                                                                   & \multicolumn{1}{l|}{\textit{1-3 times}}             & 6                                                                      & 9                                                                        \\
                                                                                                                                   & \multicolumn{1}{l|}{\textit{4-6 times}}             & 1                                                                      & 0                                                                        \\
                                                                                                                                   & \multicolumn{1}{l|}{\textit{\textgreater 6 times}}  & 1                                                                      & 1                                                                        \\ \hline
\multirow{4}{*}{\begin{tabular}[c]{@{}l@{}}\# of real RE interviews \\ conducted\end{tabular}}                                     & \multicolumn{1}{l|}{\textit{0}}                     & 7                                                                      & 7                                                                        \\
                                                                                                                                   & \multicolumn{1}{l|}{\textit{1-3 times}}             & 3                                                                      & 3                                                                        \\
                                                                                                                                   & \multicolumn{1}{l|}{\textit{4-6 times}}             & 0                                                                      & 0                                                                        \\
                                                                                                                                   & \multicolumn{1}{l|}{\textit{\textgreater 6 times}}  & 2                                                                      & 2                                                                        \\ \hline
\end{tabular}
\end{table*}
\renewcommand{\arraystretch}{1}

The participants are a mixed group of industry professionals and graduate students who are mostly the students of the ``Software Requirements Engineering” course in the Computer Engineering Department of Bogazici University. The Computer Engineering Department offers a one-year non-thesis M.Sc. degree for industry professionals besides a regular M.Sc. program in computer engineering. This course is a selective graduate course offered to both programs. The study was conducted during the 2021-2022 academic year. It was voluntary for the students to participate although the instructor encouraged them to take part in as an extracurricular class activity. The participating students received a small bonus grade.

All participants were non-native English speakers and had an age 18-24 $(n=2)$, 25-29 $(n=12)$, 30-34 $(n=5)$, 35-41 $(n=5)$. We gathered information about the participants' education and occupation, as well as their expertise and experience with software requirements elicitation interviews. Of the 24, 22 were software industry professionals (developers, testers, requirements engineers, and so on), with 17 of them also pursuing their graduate studies. We have one computer engineering undergrad student and one senior graduate student with no professional work experience. Professionals have years of working experience more than 6 years $(n=8)$, between 4-6 years $(n=8)$, between 1-3 years $(n=4)$, less than a year $(n=2)$.

Except for five, all participants had received a kind of training in software requirements engineering through university courses, self-study, and/or company training programs. Eighteen of them practiced mock interviews before one to three times $(n=15)$, four to six times $(n=1)$, more than six times $(n=2)$. Six of them had never participated in a mock interview. On the other hand, the number of participants who conducted a real interview is relatively low. Only 10 of them practiced an interview with a real stakeholder one to three times $(n=6)$, more than six times $(n=4)$.

We also gathered participants' self-evaluation scores on theoretical understanding of requirement elicitation interview techniques ($M=3, SD=0.93$) and confidence level ($M=3.12, SD=0.9$) in practicing an interview over a 5-point Likert scale (1=poor, 5=very good). The demographics of the participants are described in Table~\ref{tab:participant_demographics}.

Of 24 participants, half are assigned to \vico{} condition and the other half are assigned to \sys{} condition randomly. The experiment times were determined beforehand using an online scheduling tool according to the participants' preferences. The participant and the experimenter joined the call at the scheduled time, and the experiment was executed following the steps explained in Section \ref{sec:procedure}. The experiment lasted 11.86 minutes on average for \vico{} condition with a minimum of 8.73 minutes and a maximum of 17.37 minutes. For \sys{} condition, the participants performed the experiment in an average time of 23.72 minutes, ranging from a minimum of 17.49 minutes to a high of 30.66 minutes. The processing speed in \vico{} condition and the number of mistakes visited in the feedback phase in \sys{} condition account for most of the differences in experiment duration among the participants. The descriptive statistics for both experimental conditions' interview session duration and feedback session duration are given in Table \ref{tab:session_dur_stats}. The overall study was completed in 30 days period. 

\begin{table}[htbp]
\caption{Descriptive statistics for the duration of interview and feedback sessions in \vico{} and \sys{} conditions.}
\centering
\label{tab:session_dur_stats}
\begin{tabular}{lrrrr}
\hline
         & \multicolumn{2}{c}{\begin{tabular}[c]{@{}c@{}}Interview Session Duration \\ (in secs)\end{tabular}} & \multicolumn{2}{c}{\begin{tabular}[c]{@{}c@{}}Feedback Session Duration \\ (in secs)\end{tabular}} \\ \hline
         & M (SD)                                           & min-max                                      & M (SD)                                           & min-max                                     \\ \hline
VICO     & 636.68 (140.53)                                      & 468-965                                      & 75.19 (33.22)                                        & 28-123                                      \\
RoboREIT & 777.88 (171.06)                                      & 550-1089                                     & 645.52 (117.38)                                      & 462-821                                     \\ \hline
\end{tabular}
\end{table}

\subsubsection{Randomization Validation}
We measured the NARS scores which could strongly influence our results. We did not find a significant difference between the NARS scores of the participants assigned to the \vico{} group and the \sys{} group ($t\_stat=1.17, p=0.25$). Similarly, self-reported theoretical and practical expertise scores in conducting requirements elicitation interviews are analyzed if there was a significant difference. We used the mean of the two scores as expertise level, which did not show significance ($t\_stat=1.61, p=0.12$) between the two groups. The results reveal that the randomization was successful.

The participants in the \vico{} group had an average of 17 turns, ranging from 15 to 18. It was 17.5 with a minimum of 15, and a maximum of 19 turns for the \sys{} group. We also looked at how many mistakes were made in each scenario. In the \vico group, the participants averaged nine errors, with seven being the lowest and 11 being the highest. For the \sys{} group, the error count ranged from 7 to 13, with 9.5 being the median. Table \ref{tab:session_turn_stats} shows the descriptive statistics for the duration of the sessions, turn count, and mistake count.

\begin{table*}
\centering
\caption{Descriptive statistics for the turn count and mistake count in \vico{} and \sys{} conditions.}
\label{tab:session_turn_stats}
\begin{tabular}{@{}lrrrr@{}}
\toprule
         & \multicolumn{2}{c}{\begin{tabular}[c]{@{}c@{}}Turn  Count\end{tabular}} & \multicolumn{2}{c}{\begin{tabular}[c]{@{}c@{}}Mistake Count\end{tabular}} \\ \midrule
         & median (IQR)                           & min-max                          & median (IQR)                            & min-max                           \\ \midrule
VICO     & 17 (1)                                 & 15-18                            & 9 (1.5)                                 & 7-11                              \\
RoboREIT & 17.5 (2)                               & 15-19                            & 9.5 (3.25)                              & 7-13                              \\ \bottomrule
\end{tabular}
\end{table*}

\begin{figure*}
   \centering
	\includegraphics[scale=0.45]{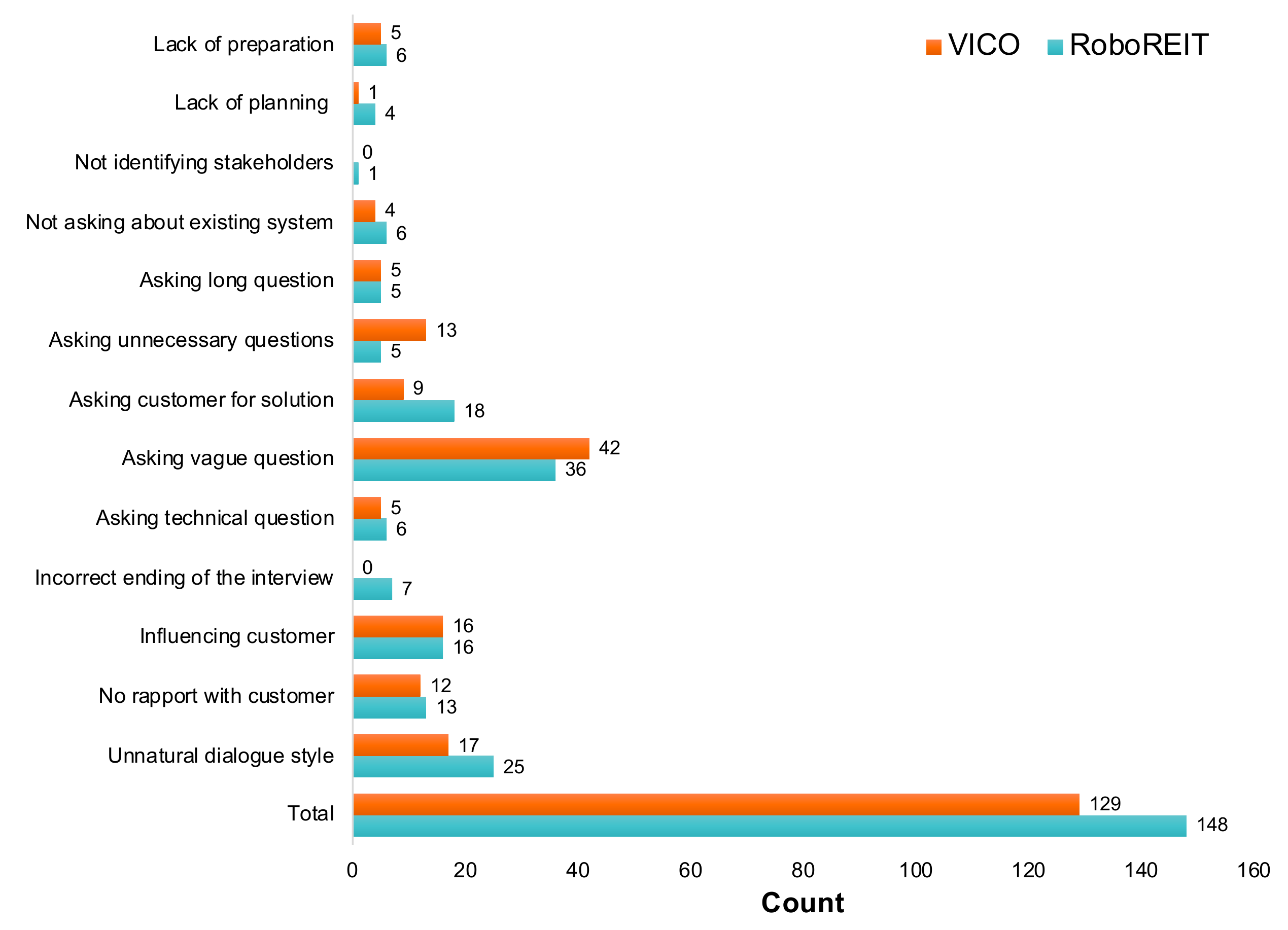}
	\caption{The number of mistakes made by the groups of \vico{} and \sys{}.}
	\label{fig:mistakes_per_condition}
\end{figure*}

We analyzed the mistakes made in terms of mistake categories as provided in~\cite{bano2019teaching}. The total number of mistakes in the \sys{} group is 148, while it is 129 for the \vico{} group. The top three most frequently made mistake classes are \textit{Asking vague question ($n=36$), unnatural dialogue style ($n=25$)}, and \textit{asking stakeholder for solution ($n=18$)} in the group of \sys{}, while they are \textit{Asking vague question ($n=42$), unnatural dialogue style ($n=17$)}, and \textit{influencing stakeholder ($n=16$)} in the group of \vico{}. The top two most frequent mistake types are the same across the conditions. The most common one, \textit{Asking vague question}, is already the most frequently induced mistake in the scenario. Figure \ref{fig:mistakes_per_condition} shows the number of mistakes per mistake class across the two conditions.

\subsection{Results}
\label{sec:results}

We test a set of hypotheses derived from our RQs. The corresponding null and alternative hypotheses for each RQ are described below, along with the statistical tests employed to examine them. We perform Shapiro-Wilk normality tests to check if the data was parametric. We analyze non-parametric data using Mann-Whitney U test \cite{nachar2008mann} and use $T$ test for parametric data \cite{cramer2016mathematical}. All hypotheses are tested for confidence level 95\% ($p \leq 0.05$). The descriptive statistics of median ($Mdn$) and interquartile range ($IQR$), and mean ($M$) and standard deviation ($SD$) are reported for non-parametric and parametric data, respectively.

Likert scales are ordinal, in five categories for our study, from 1=strongly disagree to 5=strongly agree. Although there is a rank order for the response categories, it is not possible to assume that the values' intervals are identical. The typical method is to utilize the median as the measure of central tendency because the mathematical operations necessary to derive the mean and standard deviation are inappropriate for ordinal data \cite{blaikie2003analyzing}. Hence, we apply Mann-Whitney U test for the hypotheses on the variables measured in Likert scale. This non-parametric test is appropriate for the 5-point Likert scale analysis since it has comparable power to the $T$ test for small sample sizes \cite{de2010five}.

The descriptive statistics ($Mdn$ and $IQR$) of the dependent variables for the two groups are presented in Table \ref{tab:stat_perception} along with hypothesis test statistics. In the remaining text, the participants assigned to \sys{} are referred to as Group A, while those assigned to \vico{} are as Group B.

\begin{table*}
\caption{The descriptive statistics and hypothesis test results for the dependent variables in \sys{} and \vico{} conditions.}
\label{tab:stat_perception}
\begin{tabular}{lrr|rr|r}
\hline
                                                                                             & \multicolumn{2}{c|}{RoboREIT}       & \multicolumn{2}{c|}{VICO}          & \multicolumn{1}{c}{\begin{tabular}[c]{@{}c@{}}Mann-Whitney \\ U test\end{tabular}} \\ \hline
                                                                                             & Median & IQR                        & Median & IQR                       & p-value (U stat)                                                                   \\ \hline
Processing Speed (PS)                                                                        & 7.14   & \multicolumn{1}{l|}{16.36} & 5.05   & \multicolumn{1}{c|}{1.08} & \textbf{0.003 (111.0)}                                                             \\
Attitudes (ATT)                                                                              & 4.00   & 0.70                       & 4.00   & 0.83                      & 0.50 (60.0)                                                                        \\
Perceived ease-of-use (PEU)                                                                  & 4.00   & 0.16                       & 4.67   & 0.50                      & \textbf{0.02 (33.5)}                                                               \\
Perceived usefulness (PU)                                                                    & 4.00   & 1.06                       & 4.50   & 0.56                      & 0.33 (55.0)                                                                        \\
Perceived engagament (PE)                                                                    & 3.50   & 1.00                       & 4.00   & 0.25                      & 0.08 (44.0)                                                                        \\
\begin{tabular}[c]{@{}l@{}}Perceived helpfulness\\ in finding the mistakes (PH)\end{tabular} & 4.00   & 1.25                       & 4.50   & 1.00                      & 0.78 (67.0)                                                                        \\ \hline
\end{tabular}
\end{table*}

\begin{enumerate}[start=1,label={\bfseries RQ\arabic*:}, leftmargin = 3em]
    \item \textit{How does the form of the interaction (audio-visual interaction with a physical robot in case of \sys{} and mouse click based interaction with a web agent in case of \vico{}), which the user is exposed to while conducting the interview, influence the speed of user response?}\\
    
    To answer RQ1, we consider independent samples of the processing speed variable $PS$ from group A and group B. Formally, we have $PS_A = \{PS(p_i), i=1 ... |P_A|\}$ and $PS_B = \{PS(p_i), i=1 ... |P_B|\}$. The one-tailed null hypothesis is $H_{10} = $``the processing speed in \sys{} condition is lower or equal than the one of the \vico{} condition" (i.e., $\mu_{PS_A} \leq \mu_{PS_B}$). The one-tail alternative hypothesis is $H_{11} = $``the processing speed in \sys{} condition is greater than the one of the \vico{} condition" (i.e. $\mu_{PS_A} > \mu_{PS_B}$).
    
    To assess response time $RT_t(p)$, we annotated the recorded session of participant $p$ to extract the time frames of the interview steps outlined in Section \ref{sec:system} for each turn $t$. We used ELAN\footnote{https://archive.mpi.nl/tla/elan} to annotate the video recordings for both \sys{} and \vico{} conditions. ELAN provides an easy-to-use interface where the user can add textual annotations on video and audio recordings in multiple tiers. The correctness of the participant choice for each turn was also marked as \textit{mistake} or \textit{no mistake}.
    
    \begin{figure*}
    	\includegraphics[scale=0.365]{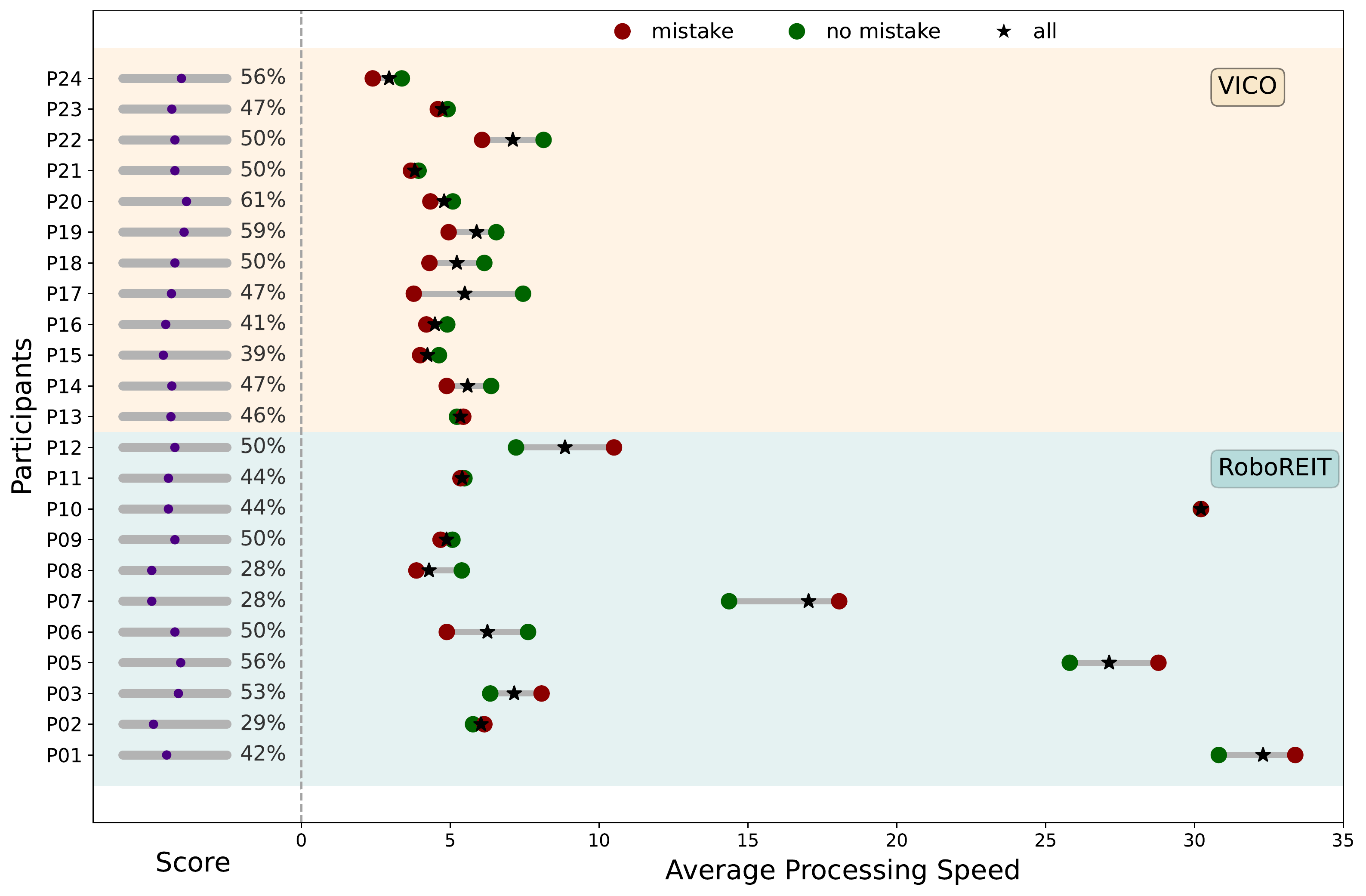}
    	\caption{Average process speed for the responses with mistake and no mistake and for all responses for the groups of \vico{} and \sys{}.}
    	\label{fig:process_speed}
    \end{figure*}
    
    As shown in Figure \ref{fig:process_speed}, the average processing speed in \sys{} condition is greater than \vico{} condition. As average processing speed variable violates the normal distribution assumption with Shapiro-Wilk's test result ($W=0.608, p < 0.001$), we applied Mann-Whitney U test to check whether average processing speed is significantly higher in \sys{} condition compared to \vico{} condition ($H_{11}$). The difference is significant, with $U=111.0, p=0.003$. Hence, $H_{10}$ is rejected and $H_{11}$ is validated.
    
    We also analyze the sub-hypothesis to focus on the average processing speed for the responses with mistake $I^{M}$ and no-mistake $I^{NM}$. Our goal is to examine the relationship between the system's interaction form and the correctness of the response with respect to the dependent variable $PS$. The average processing speed for the specific interview responses $R$ of participant $p$ is then given by averaging $PS_t(p)$ over all the turns which belong to the specific response type $I^R$, namely mistake and no-mistake:
    \begin{equation}
    PS(I^{R}(p)) = \dfrac{1}{T^R(p)} \sum_{t \in \{1...|T^R(p)|\}} {PS_t(p)} 
    \end{equation}
    
    Again, for this case, we consider independent samples of the processing speed variable for mistake and no-mistake responses from group A and group B. The one-tailed null hypothesis is defined as $H^R_{10}$``The processing speed in \sys{} condition is lower or equal than the one of the \vico{} condition for the response type $R$ where $R$ depicts the responses with \textit{mistake} or \textit{no-mistake}". One-tailed Mann-Whitney U test is performed. The difference is significant for the processing speed of responses with both mistake ($U=114.0, p=0.001$) and no-mistake ($U=103.0, p=0.012$). 
    
    \item \textit{How do \vico{} and \sys{} influence the perceived acceptance of the underlying technology in the dimensions of \textbf{RQ2a:} perceived attitudes, \textbf{RQ2b:} perceived ease-of-use, \textbf{RQ2c:} perceived usefulness}?\\
    
    To answer the sub-research questions RQ2a, RQ2b, and RQ2c, we consider independent samples of the dependent variables $ATT$, $PEU$, and $PU$, respectively, from group A and group B. Cronbach’s $\alpha$ analysis, a common method to measure the internal consistency of data, is used to evaluate the reliability of the questions in each of the surveyed dimensions based on the data of the valid samples. When $\alpha<$0.35, it means low reliability; 0.35$<\alpha<$0.70 means moderate reliability; and $\alpha>$0.70 means high reliability. 
    The reliability of the questions in each of the surveyed dimensions was higher than 0.70, indicating high reliability, particularly the questions of the dimensions of attitudes towards using (0.93), perceived ease-of-use (0.70), and perceived usefulness (0.90). The scores of each dimension's questions were added up and averaged for each participant. The average point of each dimension was then used to run the significance tests and to calculate descriptive statistics.
    
    \begin{figure*}
    \centering
    \includegraphics[scale=0.6]{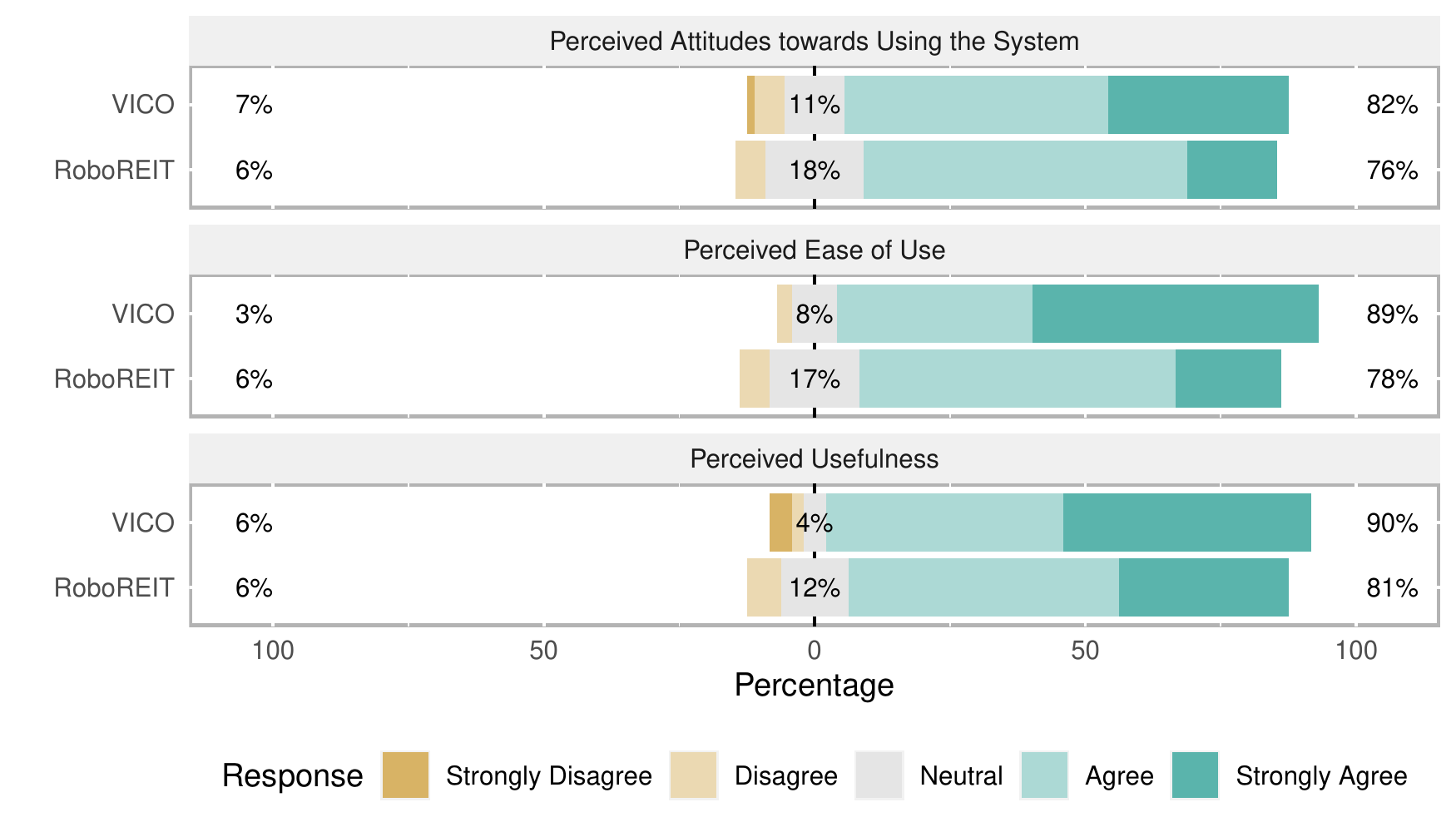}
    \caption{The questionnaire results for the technology acceptance model for the groups of \vico{} and \sys{} in the dimensions of attitudes toward using, ease-of-use, and usefulness.}
    \label{fig:p_attitudes}
    \end{figure*}
    
    \begin{itemize}
    \item[] RQ2a: Formally, we have $ATT_A = \{ATT(p_i), i=1 ... |P_A|\}$ and $ATT_B = \{ATT(p_i), i=1 ... |P_B|\}$. The two-tailed null hypothesis is $H_{2a0} = $``the perceived attitudes in \sys{} condition is equal to the one of the \vico{} condition" (i.e., $\mu_{ATT_A} = \mu_{ATT_B}$). The two-tail alternative hypothesis is $H_{2a1} = $``the perceived attitudes for \sys{} condition is not equal to the one for the \vico{} condition" (i.e. $\mu_{ATT_A} \ne \mu_{ATT_B}$). We applied two-tailed Mann-Whitney U test to test the significance. The difference is not significant ($U=60.0, p=0.50$); hence, we failed to reject $H_{2a1}$. From Figure \ref{fig:p_attitudes}, we see that a high proportion of both groups evaluated the interview experience positively. No participants expressed a strong disagreement in the \sys{} group, whereas just one participant in the \vico{} group did. 
    
    \item[] RQ2b: Formally, we have $PEU_A = \{PEU(p_i), i=1 ... |P_A|\}$ and $PEU_B = \{PEU(p_i), i=1 ... |P_B|\}$. The one-tailed null hypothesis is $H_{2b0} = $``the perceived ease-of-use in \sys{} condition is greater than equal to the one of the \vico{} condition" (i.e., $\mu_{PEU_A} \ge \mu_{PEU_B}$). The one-tail alternative hypothesis is $H_{2b1} = $``the perceived ease-of-use for \sys{} condition is lower than to the one for the \vico{} condition" (i.e. $\mu_{PEU_A} < \mu_{PEU_B}$). One-tailed Mann-Whitney U test result reveals a significant difference ($U=33.5, p=0.02$) between the two groups for perceived ease-of-use. As seen from Figure \ref{fig:p_attitudes}, the ratio of responses on the strong agreement of ease-of-use is higher in \vico{} condition compared to \sys{}. The discrepancy is understandable, though. The interaction modality of \vico{} is much simpler than \sys{}, in which the participant is required to perform active listening for the robot's speech and respond verbally. Moreover, the feedback session of \sys{} increases the participant's workload by revisiting each incorrect response and treat again. \vico{} does not force the participant to have a detailed investigation on the feedback report as the average time spent on the feedback report is just 75.19 secs as shown in Table \ref{tab:session_dur_stats}. Moreover, just two participants in \vico{} group went further in the feedback report and opened the document which describes the mistakes in detail. The simplistic interaction design and less challenging feedback component of \vico{} might result in higher scores for perceived ease-of-use compared to \sys{}.
    
    \item[] RQ2c: Formally, we have $PU_A = \{PU(p_i), i=1 ... |P_A|\}$ and $PU_B = \{PU(p_i), i=1 ... |P_B|\}$. The two-tailed null hypothesis is $H_{2c0} = $``the perceived usefulness in \sys{} condition is equal to the one of the \vico{} condition" (i.e., $\mu_{PU_A} = \mu_{PU_B}$). The two-tail alternative hypothesis is $H_{2c1} = $``the perceived usefulness for \sys{} condition is not equal to the one for the \vico{} condition" (i.e. $\mu_{PU_A} \ne \mu_{PU_B}$). Two-tailed Mann-Whitney U test shows the difference is not significant for the perceived usefulness for \sys{} and \vico{} ($U=55.0, p=0.33$). As shown in Figure \ref{fig:p_attitudes}, both group A and group B participants evaluate the system as useful with medians of 4 and 4.5, respectively. 
    \end{itemize}
    
    \item \textit{How do \vico{} and \sys{} influence the perceived engagement of training with the system?} \\
    
    To answer RQ3, we considered independent samples of the engagement variable $PE$ from group A and group B. Formally, we have $PE_A = \{PE(p_i), i=1 ... |P_A|\}$ and $PE_B = \{PE(p_i), i=1 ... |P_B|\}$. The two-tailed null hypothesis is $H_{30} = $``the perceived engagement in \sys{} condition is equal to the one of the \vico{} condition" (i.e., $\mu_{PE_A} = \mu_{PE_B}$). The two-tail alternative hypothesis is $H_{31} = $``the perceived engagement for \sys{} condition is not equal to the one for the \vico{} condition" (i.e. $\mu_{PE_A} \ne \mu_{PE_B}$). The result does not show any significance between group A and group B ($U=44.0, p=0.08$). 
    
    \begin{figure*}
    \centering
    \includegraphics[scale=0.6]{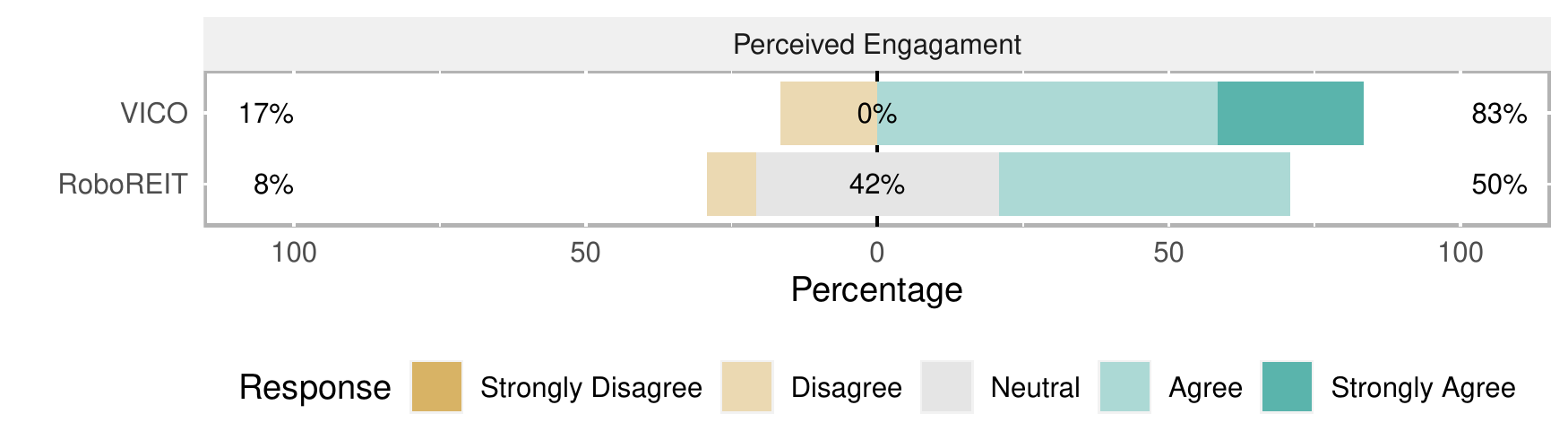}
    \caption{Perceived engagement result for the groups of \vico{} and \sys{}.}
    \label{fig:p_engagement}
    \end{figure*}
    
    \item \textit{How do the feedback components of \vico{} and \sys{} influence the perceived effectiveness and helpfulness of the system to identify the mistakes?} \\
    
    To answer RQ4, we considered independent samples of the variable of perceived helpfulness in spotting mistakes $PH$ from group A and group B. Formally, we have $PH_A = \{PH(p_i), i=1 ... |P_A|\}$ and $PH_B = \{PH(p_i), i=1 ... |P_B|\}$. The two-tailed null hypothesis is $H_{40} = $``the perceived helpfulness in spotting mistakes for \sys{} condition is equal to the one of the \vico{} condition" (i.e., $\mu_{PH_A} = \mu_{PH_B}$). The two-tail alternative hypothesis is $H_{41} = $``the perceived helpfulness in spotting mistakes for \sys{} condition is not equal to the one for the \vico{} condition" (i.e. $\mu_{PH_A} \ne \mu_{PH_B}$). The difference is not significant with two-tailed Mann-Whitney U test stats ($U=67.0, p=0.78$). From Figure \ref{fig:p_assistiveness}, we see that no participants in \sys{} group negatively evaluated the system's helpfulness. They mostly considered it effective or highly effective, and a small percentage voted neutral. In the \vico{} group, some participants did not agree that the system helps identify mistakes.

    \begin{figure*}
    \centering
    \includegraphics[scale=0.6]{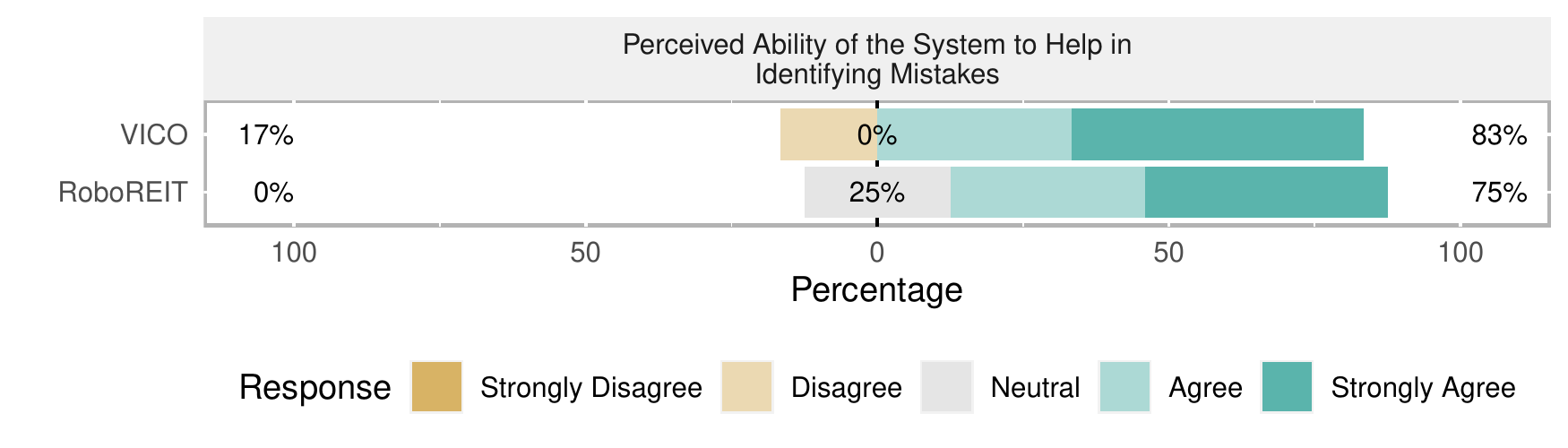}
    \caption{Perceived helpfulness result for finding the mistake for the groups of \vico{} and \sys{}.}
    \label{fig:p_assistiveness}
    \end{figure*}
    
\end{enumerate}

\subsection{Exploratory Analysis}
\label{sec:exploratory_analysis}

\paragraph{Facial Expressivity.} According to the studies on human-robot interaction, relating to a robot in a similar way that we do in human-to-human communication would improve our interactions with the robots \cite{1251796,6094592}. The feeling of being attached to a robot, and thus, the degree of interaction is influenced by many elements sourced from the perception of the robot, which is shaped by both robot's appearance, and physical and cognitive capabilities. As we perceive more human-like characteristics in the robots, we tend to interact mindlessly with them \cite{broadbent2017interactions}. In this context, anthropomorphism plays a significant role in the perception of a robot as a social agency. Humanoid robots, like Nao which is used in our \sys{} system, are thought to be more relatable. Furthermore, embodied robots evoke stronger emotional responses in people than virtual ones.

\begin{figure*}
\centering
\includegraphics[scale=0.152]{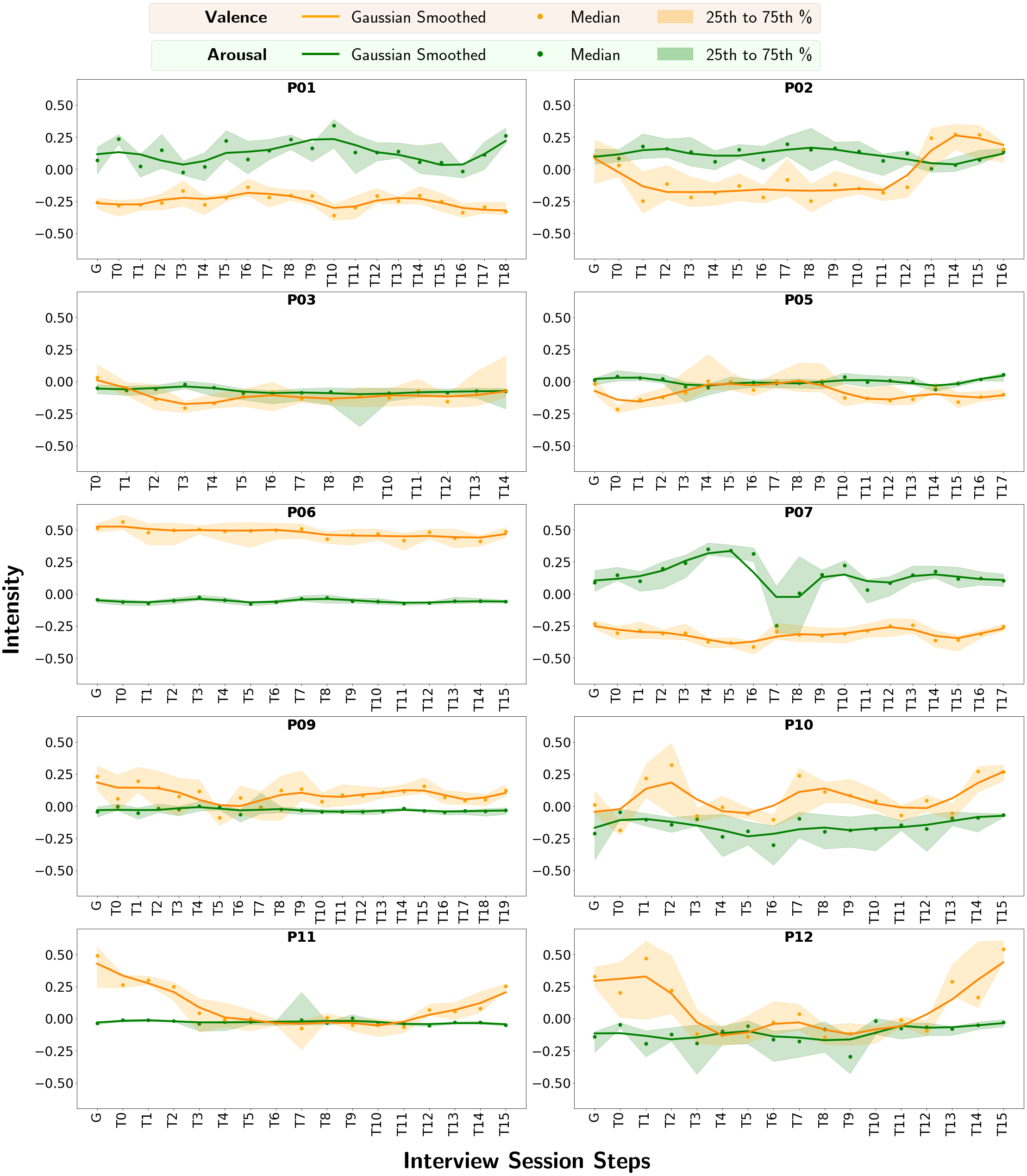}
\caption{Facial expression analysis of the participants in arousal and valence dimensions during interview sessions.}
\label{fig:exp1_fe_interview}
\end{figure*}

The face is the most effective form of communication in human social interaction \cite{konijn2020differential}. Facial expressivity plays a crucial role in communication because it allows the other person in the conversation to guess information about the affective state of the sender quickly. In an educational setup like our interview trainer system, the affects that emerged during a learning process belong to a specific subset of affects like frustration, boredom, confusion, and enjoyment~\cite{zhang2022educational} by which the learner's cognitive situation is conveyed to the teacher. Monitoring the learner's facial expressions would help shape the course of the training by adjusting the pace of teaching or balancing the intensity of the provided content. 

Monitoring facial expressions could also be helpful while conducting requirements elicitation interviews. Students should prepare not just for technical knowledge but also for soft skills like interview management and building rapport with the stakeholder in order to become an expert at interviewing. Observing and evaluating the participants' facial expressions might allow feedback on their social responses and help them improve their soft skills. 

Having all these in mind, we performed an exploratory analysis on the facial expressivity of the participants in the condition of \sys{} where the audio-visual interaction was performed with an embodied robot. Automatic analysis of affect in the face images, using open source library FaceChannel \cite{barros2020facechannel}, is carried out at participant level along arousal and valence dimensions. Frames are extracted from a given video segment and cropped to contain the face region. The extracted face images are then fed into a lightweight deep neural network model provided by FaceChannel to produce arousal and valence values in the range [-1, 1]. To have descriptive statistics of the values for a given interval, the median and inter-quartile range ($25^{th}$ to $75^{th}$ percentile) are calculated. Additionally, the median points are filtered by a Gaussian smoothing kernel with parameters window size $M=3$ and standard deviation $\sigma=1$ to visualize the overall trend. We did the study separately for the interview and feedback sessions since the activity carried out in each session can impact the participants' facial expressivity. During the interview session, participants take the role of a requirements engineer in a requirements elicitation interview, while they turn into trainees in the feedback session receiving instruction from \sys{} on their mistakes. P04 and P08 are not considered in this analysis because of faulty recording and poor illumination, respectively.

For the interview session, the arousal and valence analysis is performed for the greeting phase and each turn per participant. As depicted in Figure \ref{fig:flow}, an interview turn has three steps; consideration of the options ($T3$), responding with selected option ($T4$), and receiving response back from \sys{} ($T5$) (see Section \ref{sec:system} for details). We exclude the interval of $T4$ step as the mouth area movements related to language articulation during talking would confound the detection of arousal and valence values. As shown in Figure \ref{fig:exp1_fe_interview}, participants P03, P05, P06, P09, and P11 expressed highly stable arousal levels throughout the interview session. P01, P02, P10, and P12 showed mild deviations over the turns, but their interquartile range within some turns is slightly high. The arousal level of P07 exhibits a considerable decline in the midst of the interview before stabilizing once more in the closing stages. In valence analysis, we observe a pretty stable pattern for participants P01, P03, P06, and P07. Likewise, P05 and P09 expressed consistent valence expressions with slightly higher variances at a few of the turns, which could point out mild frowning or smiling. Early in the interview, P10 experiences large positive spikes but then shows lower deviations. P11 and P12 begin the interview with highly positive expressions, shift to neutral in the middle, and then make positive expressions again towards the end.

\begin{figure*}
\centering
\includegraphics[scale=0.152]{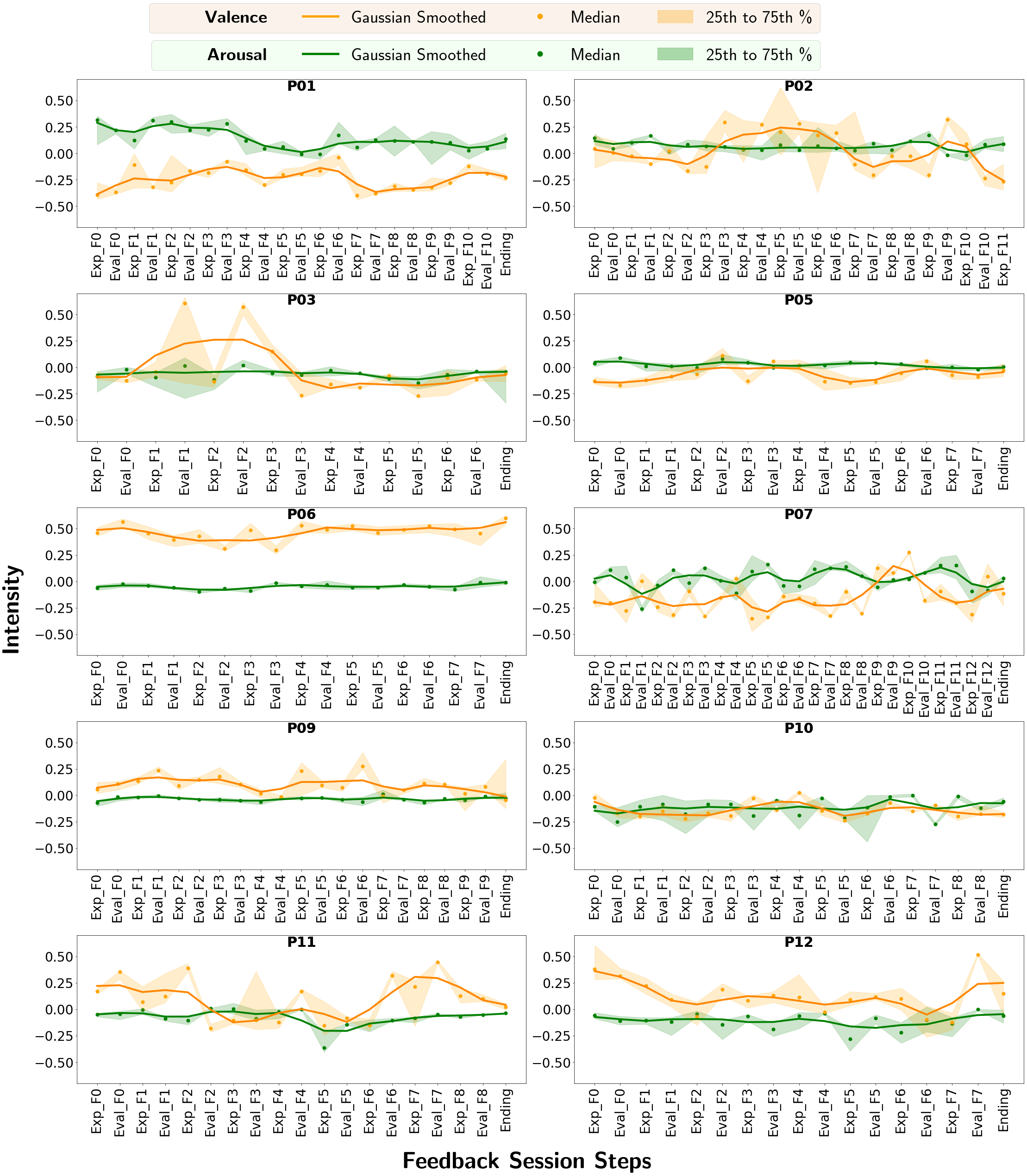}
\caption{Facial expression analysis of the participants in arousal and valence dimensions during feedback sessions.}
\label{fig:exp1_fe_feedback}
\end{figure*}

For the feedback session, we presented the descriptive statistics for arousal and valence values for each revisited turn of a participant and for the ending phase, as shown in Figure \ref{fig:exp1_fe_feedback}. Each feedback turn has three steps; presentation of the incorrect option and accompanying feedback ($T7$ and $T8$), re-consideration and response of the participant ($T9$), and re-evaluation and confirmation of the accuracy of the chosen option ($T10$ and $T11$) (see Section \ref{sec:system} for the details). We excluded step $T9$ from monitoring facial expressions, for the same reason as disregarding step $T4$. Steps $T7$, $T8$, and $T9$ were observed separately because we wanted to see how participants reacted when they were notified about their mistakes ($Exp\_F$) and succeeded or failed on the second attempt ($Eval\_F$). Participants P05, P06, and P09 displayed relatively stable expressions in both the valence and arousal dimensions throughout the feedback session, just like they had during the interview. Interestingly, P10 displayed a similar pattern to them even though she was more expressive while conducting the interview. The arousal levels of P02, P03, P11, and P12 remain nearly constant throughout the session. On the other hand, in their valence values, we detect peaks, particularly for some of the $Eval\_F$ steps. For example, P03 has two significant positive spikes in the steps of $Eval\_F1$ and $Eval\_F2$, as a reaction to \sys{}'s acknowledgment of the accuracy on the second attempt. For P01 and P07, we observe a more divergent pattern for both dimensions over the session, but the divergences within each step are relatively small. 

\begin{figure}[htbp]
   \centering
	\includegraphics[scale=0.56]{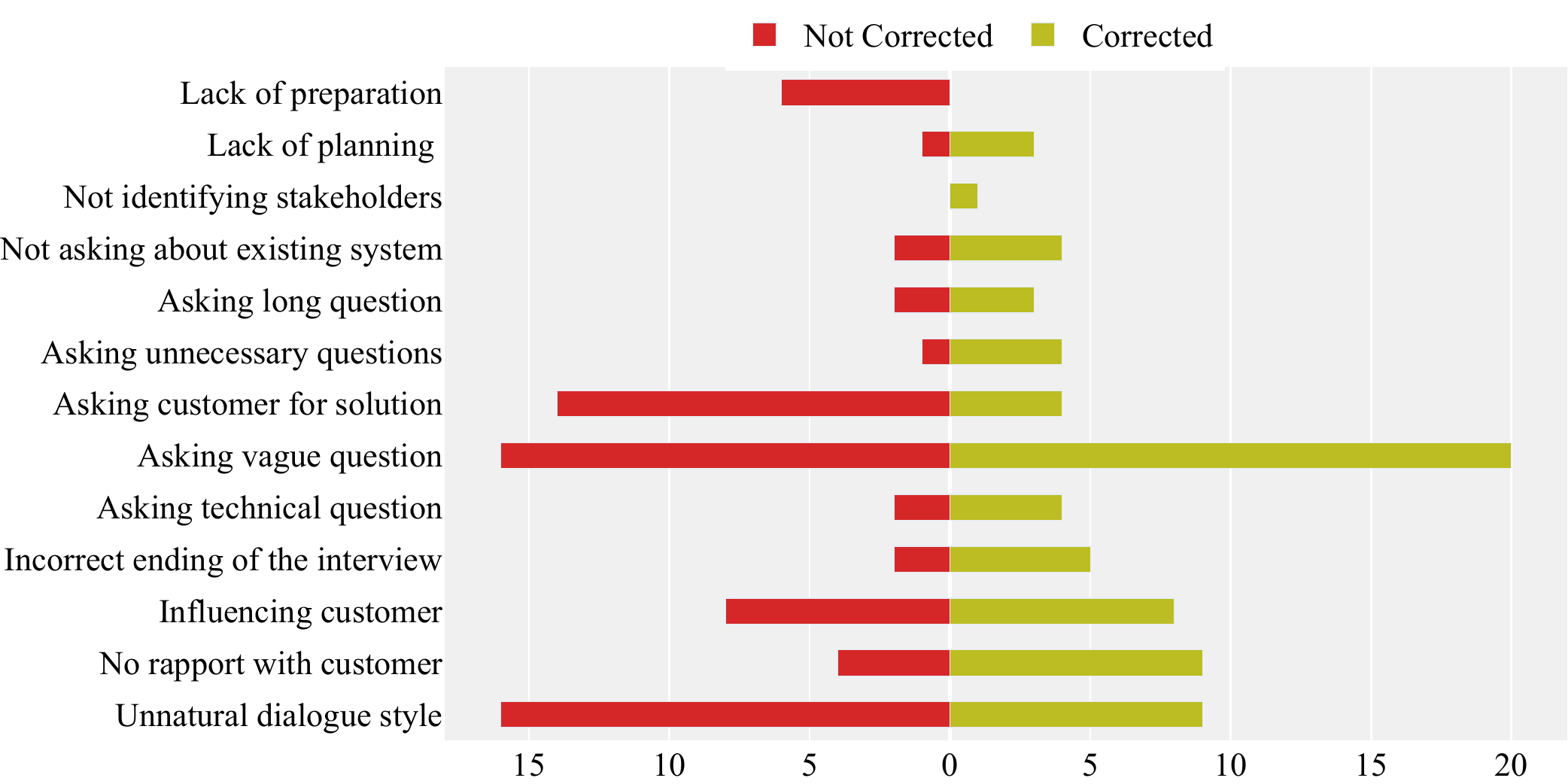}
	\caption{Following feedback, total corrected and uncorrected mistake counts for each mistake group.}
	\label{fig:correction_ratio_after_feedback}
\end{figure}

\paragraph{Feedback effectiveness.} In the condition of \sys{}, participants were provided feedback for each of their incorrect choices and allowed to retry. We checked the total number of corrected and uncorrected mistakes for each mistake category to determine how well participants could utilize the feedback and were able to fix their faults. If a participant picked the right response after receiving feedback for their error, we marked the mistake as \textit{corrected}. If the selected option still contains some associated errors, we classified the initial mistake as \textit{uncorrected}. Figure \ref{fig:correction_ratio_after_feedback} shows the counts of corrected and uncorrected mistakes per mistake category. Out of 148 mistakes in total, 74 of them are corrected. Among all the mistake groups, \textit{lack of preparation, asking stakeholder for solution} and \textit{unnatural dialog style} seem to be the most challenging ones where the ratio of corrected count over the total count is below 0.36 for them. The correction ratio is higher than 0.5 for every other mistake group. Although more research is required to fully grasp the factors influencing participants' ability to correct mistakes, we argue that the quality of the feedback is crucial. 

\section{Threats to Validity}
\label{sec:threats_to_validity}
In this section, we discuss the existing threats to our system and, when possible, how they have been mitigated. The main threats, based on Wohlin \textit{et. al.} \cite{wohlin2012experimentation} were as follows:

\paragraph{Internal validity.} Biased attitudes towards robots may affect our results, especially for research questions RQ2, RQ3, and RQ4. To address this issue related to the possible selection bias between the experimental groups, we checked if the groups had significantly different attitudes toward the robots. We used a widely applied questionnaire to measure the negative attitudes towards robots~\cite{syrdal2009negative} and applied a statistical test to show there is no significant difference between the two experimental groups so that it can not constitute as a confounding factor for our results. Likewise, we checked the perceived expertise in conducting requirements elicitation interviews to ensure it is balanced across the groups. Again, we did not find a significant difference. The details for the randomization validation is provided in Section~\ref{sec:study_execution}.

To mitigate the possible social threats to internal validity, we carefully designed the introduction of the experiment to the participants. The experimental condition is not mentioned to them but just the high-level context and aim of the study by both the video and written document to mitigate possible resentful demoralization and compensatory un-equalizationalization of treatments. Participation was entirely voluntary although the experiment was encouraged by the instructor and proposed as an external exercise within the class and rewarded with a small bonus for the course grade. Student participants were not graded based on their performance in the interview, and this was explicitly communicated to them before the experiment. Likewise, other participants were not provided any rewards or compensation for participation in the experiment. 

\paragraph{External validity.} Our heterogeneous participant group comprises graduate students of a specific course of requirements engineering and of industry professionals who are required to use requirements elicitation practices at some phases in their jobs. Their years of work experience and prior involvement in requirements elicitation interviews also differ. Although having such a mixed group of participants induces a threat to conclusion validity, we prioritize external validity over conclusion validity for our study, as suggested in \cite{wohlin2012experimentation}. Since, for empirical software engineering research, it is highly important to show the applicability of a study for a broader user population. We can argue that our sample population is a good representative of the target user group of the experiment, which is determined to be anyone from any level of expertise pursuing the purpose of excelling in conducting requirements elicitation interviews. With this, we might potentially lessen the validity concern presented by the small number of participants, as Falessi \textit{et al.}~\cite{falessi2018empirical} reported that a small number of participants could be considered sufficient for producing credible results when they accurately reflect the target user population. 
Moreover, the system used in our experiment is developed in a way that is ready to deploy without requiring any further significant developments. By doing so, we aim to mitigate the threat which might be induced by an experimental setting that is not representative of real-world practice.

\paragraph{Construct validity.} People often perform better when a learning experience is new. Short-term studies, which are not repeated a number of times, do not address the novelty effect that the participants' feeling of being new to the exposed experience might boost the results for a short period. Like most user studies that were performed in requirement engineering education, we could not address this problem in our study due to the minimum availability of the participants. Nonetheless, we designed the experiment as between-subject to minimize the learning effect across the trials. Each participant experienced just one condition. Evaluation apprehension might induce a social threat to construct validity of our study. When performing in front of others, people may become anxious about being watched and judged, which may impact the study's findings. To mitigate this threat, the experimenter was not visible to the participants throughout the study.

\paragraph{Conclusion validity.} Consistency and verifiability of the results are key components of the reliability of research. Despite the fact that any user study investigation is relatively subjective and does not guarantee 100\% of the repeatability of the results, we tried to mitigate this problem by employing widely adopted and well-studied instruments to measure the participants' opinions for our research objectives. To ensure the reliability of treatment implementation, we defined the experimental procedure during the study design and applied the same flow to every participant. Since the systems used in the experiments are designed to be autonomous, there is no possibility of researcher bias during the execution of the experiments except for wizarding speech-to-text functionality. However, since participant speech is taken as it is and accepted only if it meets one of the available dialogue options, there is no room for the researcher to induce any application bias. In the experiment introduction, we explicitly defined the required environment to minimize the effect of irrelevant components in the experimental setting, like background noise or the presence of others which might distract the participant during the experiment. We also defined the technical requirements like connecting on a computer that has a wired internet connection using high-quality headphones in a well-illuminated room. However, since the experiment was held online, we could not entirely control the participants' experimental environments. 

\section{Conclusions and Outlook}
\label{sec:conclusion}

This paper introduces \sys{}, a robotic interview trainer system designed for requirements elicitation interview education and training to help learners excel in conducting interviews. Our research is substantially novel for there is no other study published within the RE community considering the use of robots in requirements engineering education and training. We share the design of the overall system and the implementation details in~\cite{binnur_gorer_2022_7263541} and call upon the community to further investigate and improve on our work. We evaluated \sys{} with a sample of prospective users in comparison with \vico{}, a web-based interview simulator~\cite{debnath2020designing}. We quantitatively assessed \sys{} by inspecting the participants' interactions with the system and their responses to a series of questionnaires.

The participants respond significantly faster in \sys{} than in \vico{} during interview turns. This may be due to the audio-visual interaction modality provided by an embodied social robot, the participants are nudged to behave like in human-human conversation. This supports our design goal of developing a more realistic interview setup. \sys{} is rated favorably by the participants (i.e., higher than 3 = moderate level), and  we do not detect a significant difference in the perceived acceptance of \sys{} over \vico{}. \vico{} received higher ratings for perceived ease-of-use than \sys{}, which makes sense considering its more basic design. \vico{} has a simplistic web-based interface in which the user can conduct the interview by reading the texts on the screen and responding with a mouse click where \sys{} challenges the participant to listen to the robot actively and verbally responding by the selected option. Moreover, the feedback component of \sys{} is more demanding because it requires revisiting each incorrect interview turn by obtaining feedback and performing a second examination. Participants found \sys{} engaging and helpful in identifying the mistakes, although we can not observe a significant difference with the corresponding rating of \vico{}. 

\subsection{Discussion}
\label{sec:discussions}

There is a growing interest in the literature to utilize automated tools in helping learners to improve their social skills in the given context (e.g., job interviews, vocational training)~\cite{gebhard2018serious,baka2022social,othlinghaus2020technical,zheng2021serious}. In the domain of requirements elicitation interviews, soft skills also play an essential role in obtaining comprehensive needs from a stakeholder and should be included in REET activities. It may be possible to train students in social skills by monitoring and delivering feedback on the automated identification of their emotions and social attitudes in real-time through facial expressions, voice, or gestures during interaction with the interview system. However, for users to engage in expressive activities, they must first view the system as social~\cite{ekman1994nature}. Our analysis of the participants' facial expressions revealed that they reacted to the robot on various social levels, supporting our claim that \sys{} was viewed as a social agent by the participants. Even though more research is required on this topic, enhancing \sys{} with automated expression detection techniques may aid participants' soft skill development.

\emph{Qualitative remarks. }We asked the participants their most and least favorite aspects of \sys{} and \vico{}, and their suggestions for improvements as open-ended questions. The participants liked the simplicity of \vico{} and complained about the design of the dialog options. Although some participants appreciated the multi-choice style of \vico{}, others found it rather constraining their communication during the interview. The participants found some of the choices to be too similar to differentiate, and listed long sentences as a drawback as well. The participants also complained about the mistake report generated by \vico{}. The suggestions include improvements to the dialog options. 

Even though \sys{} uses the identical scenario and dialog options, the participants did not primarily discuss these and focused their comments on the robotic component. They frequently noted that the robot's text-to-speech module could be enhanced, along with the voice quality. Since the participants were non-native English speakers, language barriers may also be an issue. Since the sessions were held through a video conferencing tool, the connection quality may also have an impact on the voice quality. Nonetheless, the participants' comments show that technical excellence is highly desired for such a high-technology platform. The participants generally mentioned that they appreciated the opportunity to converse with a real robot. One participant stated their satisfaction exactly like \textit{``It is a great idea to conduct elicitation interviews with a robot, and it can help practice rehearsal interviews"}. The following was highlighted by a participant as a system drawback:\textit{``It felt that the robot lacked human emotions and responses."}, which indicates how high the participant's expectations can be for the robot's social skills. The feedback component of \sys{} is frequently observed as being highly valued, as evidenced by the statements such as \textit{``I liked most that I received feedback based on my choices, and I was able to improve myself."} and \textit{``I truly liked talking about the mistakes and learning from them."}. Some participants suggested that allowing them to ask more questions about their mistakes would be an improvement to the feedback utility. In addition, one participant proposed that if the second attempt yielded an inaccurate result, more explanations should be provided. In conclusion, participants commented more about interaction-related features and more sophisticated requests for \sys{} than they did for \vico{}.

\paragraph{Limitations of the system. }The limited dialogue options restrict users' flexibility to compose responses in their own words based on what is on their minds. This could reduce training effectiveness as the interview context will be controlled by the previously created dialogue structure. With the growth of AI technologies, this constraint might be overcome by utilizing intelligent chatbots which can interpret the user's response and reply appropriately. However, the language used in the requirements elicitation interviews has domain-based specific requirements that might not be met with a system designed for colloquial language patterns. In the future, we aim to create a customized chatbot that can identify keywords and patterns relevant to a particular scenario, helping to generate the correct answer to the user's query. Employing such a chatbot could significantly improve \sys{} in terms of usefulness and usability. However, more research will be required to ascertain the actual impact through long-term user studies.

\paragraph{Challenges. } It was challenging to plan and carry out a remote experiment, especially when the participation was voluntary. Due to the difficulty of organizing lengthier experiment sessions and the risk of fatigue and learning carryover when utilizing the same scenario repeatedly, we conducted a user study employing a between-subject design. Although we validated randomization in the experimental groups for negative attitudes toward robots and demographics, there might still be some individual differences between the two groups that could affect the results regarding the perception of the system. Designing a within-subject study might help to boost the validity of the comparative results.

\paragraph{Implications on REET.} Our system and the findings from its empirical evaluation have produced new perspectives for REET research in terms of utilizing social robots as an educational tool. The system is evaluated with participants, including the students of a requirements engineering course in the graduate level, as their extra curriculum activity. We demonstrated the applicability and usability of the system in a real class setting. Following the online learning trends that emerged during the Covid-19 pandemic, we have built our system to be used remotely via a video conference tool at any time, from anywhere, without requiring the user to be physically there with the robot, though it can function in person if preferred. Even in the remote settings, the participants could use and interact with the system successfully. The system is designed to function autonomously (except for speech to text component) to minimize human intervention. Furthermore, it is designed to work with any other scenario prepared with different use cases or a specific subset of mistake types. With all of these, we argue that educators might enable \sys{} as a supplementary tool in their REET courses by requiring the students to utilize the system often to develop their interviewing abilities.

\subsection{Future Work}
\label{sec:FutureWork}
Recent developments in deep neural networks have made it possible for publicly accessible automatic speech recognition systems to tackle the text-to-speech problem with high accuracy \cite{radfordrobust}. As the next step, we would like to improve \sys{} by replacing its human-operated speech-to-text component with one of the speech recognition libraries, enabling the entire system to function autonomously. In this way, users can access \sys{} at any desired time with a reservation system without any need to contact a human operator. Although utilizing a physically embodied robot has numerous positive social effects and enhances the user experience, it also limits the number of users that can be served simultaneously. The immersive technologies, such as virtual and augmented reality, seem promising in transferring our physical world interactions to a 3D virtual world with the help of specialized multi-sensory equipment such as VR headsets~\cite{slater2016enhancing}, which are more likely to be accessible compared to a physical robot. Metaverse, which is expected to feature these technologies in order to resemble the real world in terms of space and time, holds tremendous potential for several areas including education~\cite{metaverse}. Parallel to these advancements, it would be quite interesting to move \sys{} to a 3D environment by building a virtual reality avatar of our robot and doing the training there to assess user experiences.

In contrast to a one-size-fits-all strategy, adaptive systems play a significant role in education by offering the proper context in a personalized setting to enhance the learning outcome of the learner~\cite{ahmad2017systematic}. Our system can be augmented with different scenarios where the challenge of the context or the induced mistake types can be adjusted according to the expertise level of the learner. In this way, we could improve the learning outcome by avoiding boring experienced learners with simple training or demotivating inexperienced learners with challenging content. Moreover, we would like to develop an automatic facial expression and gesture analysis module within \sys{} to actively monitor the user during the interview session and provide feedback on behavioral skills. This could also allow \sys{} to adjust its social behaviors in the stakeholder role in order to challenge the users more on their weaker skills.

\section{Ethics Statement}
The studies involving human participants were reviewed and approved by the Ethics Committee of Bogazici University. The participants provided their informed consent to participate in this study.

\section{Conflicts of Interest}
The authors declare that the research was conducted in the absence of any commercial or financial relationships that could be construed as a potential conflict of interest.



\bibliographystyle{spmpsci}  
\bibliography{gorer}
\end{document}